\documentclass[journal]{IEEEtran}
\usepackage{amsmath}
\usepackage{amsfonts}
\usepackage{array}
\usepackage[caption=false,font=normalsize,labelfont=sf,textfont=sf]{subfig}
\usepackage{textcomp}
\usepackage{stfloats}
\usepackage{url}
\usepackage{verbatim}
\usepackage{graphicx}
\usepackage{cite}
\usepackage{multirow}
\usepackage{multicol}
\usepackage{changepage}
\usepackage{booktabs}
\usepackage{tabularx}
\usepackage{adjustbox}
\usepackage{booktabs}
\usepackage{footnote}
\usepackage{subfloat}
\usepackage{amssymb}
\usepackage{float}
\usepackage{threeparttable}
\usepackage[ruled,linesnumbered]{algorithm2e}
\usepackage{algpseudocode}
\usepackage{flushend}
\usepackage{cite}
\usepackage{color}
\usepackage{booktabs}
\usepackage[table]{xcolor}
\usepackage[citecolor=blue]{hyperref}
\begin{document}

\title{CMF-IoU:  Multi-Stage Cross-Modal Fusion 3D Object Detection with IoU Joint Prediction}

\author{Zhiwei Ning, Zhaojiang Liu, Xuanang Gao, Yifan Zuo*, Jie Yang*, Yuming Fang, Wei Liu*

\thanks{Zhiwei Ning, Zhaojiang Liu, Xuanang Gao are with the School of Automation and Intelligent Sensing \& Institute of Image Processing and Pattern Recognition, Shanghai Jiao Tong University, Shanghai, China (e-mail: zwning@sjtu.edu.cn; zhaojiangliu@sjtu.edu.cn; fangkuar@sjtu.edu.cn)}

\thanks{Yifan Zuo and Yuming Fang are with the School of Computing and Artificial Intelligence, Jiangxi University of Finance and Economics, Nanchang, Jiangxi 330032, China (e-mail: kenny0410@126.com; fa0001ng@e.ntu.edu.sg).}

\thanks{Jie Yang and Wei Liu are with the School of Automation and Intelligent Sensing \& Institute of Image Processing and Pattern Recognition \& Institute of Medical Robotics, Shanghai Jiao Tong University, Shanghai, China (e-mail: jieyang@sjtu.edu.cn; weiliucv@sjtu.edu.cn)}

\thanks{This work is partially supported by National Natural Science Foundation of China (Grant No. 62376153, 62402318, 24Z990200676, 62271237, U24A20220, 62132006, 62311530101) and Science Foundation of the Jiangxi Province of China (Grant No. 20242BAB26014).}

\thanks{Wei Liu, Jie Yang and Yifan Zuo are the corresponding authors of this paper.}
}
\markboth{IEEE TRANSACTIONS ON CIRCUITS AND SYSTEMS FOR VIDEO TECHNOLOGY, August 2025 }
{Shell \MakeLowercase{\textit{et al.}}: A Sample Article Using IEEEtran.cls for IEEE Journals}

\maketitle

\begin{abstract}
Multi-modal methods based on camera and LiDAR sensors have garnered significant attention in the field of 3D detection. However, many prevalent works focus on single or partial stage fusion, leading to insufficient feature extraction and suboptimal performance. In this paper, we introduce a multi-stage cross-modal fusion 3D detection framework, termed CMF-IOU, to effectively address the challenge of aligning 3D spatial and 2D semantic information. Specifically, we first project the pixel information into 3D space via a depth completion network to get the pseudo points, which unifies the representation of the LiDAR and camera information. Then, a bilateral cross-view enhancement 3D backbone is designed to encode LiDAR points and pseudo points. The first sparse-to-distant (S2D) branch utilizes an encoder-decoder structure to reinforce the representation of sparse LiDAR points. The second residual view consistency (ResVC) branch is proposed to mitigate the influence of inaccurate pseudo points via both the 3D and 2D convolution processes. Subsequently, we introduce an iterative voxel-point aware fine grained pooling module, which captures the spatial information from LiDAR points and textural information from pseudo points in the proposal refinement stage. To achieve more precise refinement during iteration, an intersection over union (IoU) joint prediction branch integrated with a novel proposals generation technique is designed to preserve the bounding boxes with both high IoU and classification scores. Extensive experiments show the superior performance of our method on the KITTI, nuScenes and Waymo datasets. The code is available at \url{https://github.com/pami-zwning/CMF-IOU}.
%Code will be publicly available at \url{https://github.com/zwning/CMF-IoU}.
\end{abstract}
%Additionally, the corresponding RGB values and pixel coordinates of pseudo points are temporarily reserved to avoid deviation from unnecessary projection in subsequent stages.

% uniform format: point couds
\section{Introduction}

\IEEEPARstart{O}{bject} detection in 3D real-world is regarded as an essential tasks in autonomous driving \cite{mao20233d}. Benefiting from the high-precision LiDAR sensors, numerous methods based on point clouds have been proposed \cite{qi2017pointnet,qi2017pointnet++,liang2024pointmamba}.  LiDAR information is processed by three main approaches, including point set abstraction in PointNet \cite{qi2017pointnet}, voxelization followed by 3D convolution in VoxelNet \cite{zhou2018voxelnet}, and voxel-point interaction in PV-RCNN \cite{shi2020pv}. However, current LiDAR sensors typically struggle with sparse point clouds and lack textural information compared with the camera. As a lower-cost alternative, the camera can provide 2D images with dense textural information for object detection. However, image-only methods \cite{chen2017multi,li2022bevformer,wang2022detr3d} often suffer from inevitable depth bias without the assistance of 3D localization in LiDAR sensors. To effectively combine these two complementary data sources, many multi-modal approaches \cite{liu2023bevfusion, liang2022bevfusion, wu2023virtual, bai2022transfusion, yan2023cross} are proposed and achieve better performance than single-modal methods.

Although multi-modal methods show significant potential, unbiased and comprehensive fusion of heterogeneous modalities remains a major challenge. To address the problem, previous works apply various strategies to eliminate the misalignment and improve fusion effectiveness. For example, PointPainting \cite{vora2020pointpainting} utilizes the image segmentation results to enhance point clouds in the early stage, while EPNet \cite{huang2020epnet} incorporates the encoded image features into voxel features in the middle stage. Both approaches employ an additional backbone for image representation. However, the independent encoding for image and point data fails to address the dimensional misalignment effectively. Other methods like MVP \cite{yin2021multimodal} and SFD \cite{wu2022sparse} project the image into 3D space for an unified representation. However, they focus on fusion in either the early or the late stage, which is usually insufficient and suboptimal.  Therefore, it is essential to adopt specific fusion modules across all the pipeline stages.

Moreover, most two-stage detectors \cite{shi2020pv ,deng2021voxel,li2022spatial} rely on non-maximum suppression (NMS) to retain the proposals with higher classification scores during the post-processing. However, the evaluation metric is primarily determined by the IoU values between the predicted and ground-truth boxes, rather than the classification scores. Therefore, some methods \cite{li20203d, zhu2021iou} incorporate an IoU prediction branch into the refinement head and filter out redundant boxes based solely on the predicted IoU scores. Nevertheless, the IoU values of bounding boxes generated from the regional proposals network (RPN) typically exhibit a long-tailed distribution, which poses challenges for training convergence. Therefore, it is necessary to balance the IoU and classification scores and to design a more reasonable proposal generation strategy.
 
To address the limitations of previous methods in integrating LiDAR and camera information, we propose CMF-IoU, a novel multi-stage cross-modal fusion 3D detection framework with IoU joint prediction. Unlike previous methods that process camera and LiDAR data by a 2D encoder and a 3D encoder, \cite{vora2020pointpainting, liu2023bevfusion,huang2020epnet}, we transform the pixel values from the image into the 3D space via a depth completion network in the early stage. This ensures a unified representation of depth-inferred image points (denoted as ``pseudo points'') and LiDAR points (denoted as ``raw points''). However, depth estimation inevitably introduces bias, which is typically non-Gaussian and tends to concentrate around object boundaries \cite{wu2023virtual}. Meanwhile, the raw points of the foreground objects are usually sparse and incomplete, resulting in the absence of structural representation. To address these issues, we propose a bilateral cross-view enhancement backbone to encode the above two point clouds separately, rather than directly merging them as \cite{vora2020pointpainting, yin2021multimodal}. The first residual view consistency (ResVC) branch incorporates both 3D and 2D convolutions to alleviate the impact of depth bias in pseudo points. The second sparse-to-distant (S2D) branch utilizes an encoder-decoder architecture to enlarge the receptive field for the sparse raw points. Benefiting from the spatial alignment between voxel features of raw points and pseudo points, we can directly concatenate their channel dimension for middle stage fusion, thereby avoiding the less flexible 2D-to-3D feature projection used in \cite{huang2020epnet}. In the refinement stage, an iterative voxel-point aware fine grained pooling module is designed to progressively optimize proposal locations and sizes. At each iteration, voxel and point features within the proposals are aggregated to predict regression targets, similar to \cite{shi2020pv}. Additionally, we introduce a cross-attention mechanism to enable interaction between proposals across different iterations for better refinement. 

Furthermore, for more reasonable NMS post-processing, we design an IoU-classification balanced metric rather than the single metric used in \cite{li20203d, zhu2021iou,deng2021voxel}. In detail, we add an IoU prediction branch into the original detection head to reserve the predicted boxes with both high IoU and classification scores. To ensure stable training convergence, additional proposals are generated from the ground truth (GT) boxes and combined with the proposals predicted by RPN to achieve a more uniform IoU distribution in the training stage. In summary, our contributions are as follows: 
\begin{itemize}
    \item We propose a novel 3D detector, termed CMF-IOU, which adopts a multi-stage cross-modal fusion strategy throughout the entire pipeline and integrates an IoU prediction head to provide a more suitable metric for the proposal selection.
    \item An effective bilateral cross-view enhancement backbone is designed for better feature extraction of both the pseudo points and raw points, which alleviates the impact of noise in the pseudo points via 3D-to-2D convolution and enhances long-range feature interaction in the sparse raw points via an encoder-decoder structure.
    \item We introduce an iterative voxel-point aware fine grained pooling module in the refinement stage, which enables the model to aggregate both the spatial and textural information within proposals.
    \item An IoU-classification balanced prediction metric is proposed to supervise the proposals generated from the GT bounding boxes and the RPN during the training and to facilitate a more accurate NMS during the inference.
\end{itemize}

\begin{figure*}[!ht] %%图1
	\centering  %插入的图片居中表示
	\includegraphics[width=\textwidth]{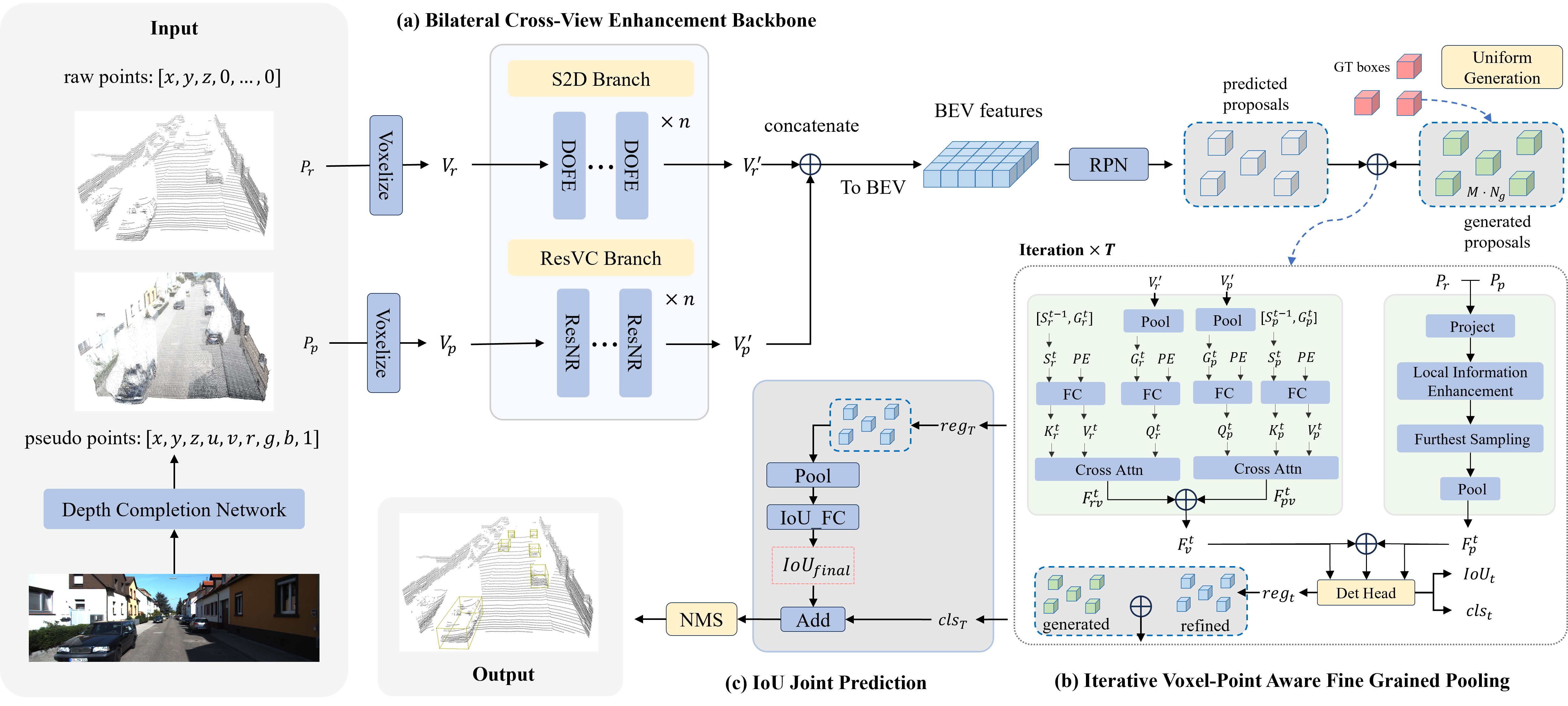} 
	\caption{Overview of our CMF-IOU framework. We estimate the pseudo points by a depth completion network and integrate them with the LiDAR points as our input. (a) The bilateral cross-view enhancement backbone contains the S2D branch and the ResVC branch, where the S2D branch encodes the raw voxel features and the ResVC branch encodes the pseudo voxel features. (b) The iterative voxel-point aware fine grained pooling is designed for optimizing the predicted and generated proposals. (c) The IoU joint prediction balances the IoU and classification scores in the NMS post-processing stage.}
	\label{fig:framework}  
\end{figure*}

\section{Related Work}

\subsection{Single-Modal 3D Detection Methods}

Most previous single-modal 3D detection methods focus on either point clouds or multi-view images, leveraging the inherent consistency within a single modality for representation learning. Considering the intuitive and structured information in LiDAR points, previous methods \cite{qi2017pointnet, qiao2024local, qi2017pointnet++, shi2019pointrcnn, tian2023medoidsformer, zhou2018voxelnet, shi2020points, sheng2021improving, wang2023long} are proposed to extract features of the 3D scene. As a milestone in the 3D task, the set abstraction proposed in PointNet \cite{qi2017pointnet} is broadly utilized in the subsequent methods. PV-RCNN \cite{shi2020pv} and CT3D \cite{sheng2021improving} are seminal two-stage works for high-performance proposal refinement. However, limited by the sparsity and irregularity of LiDAR point clouds, detailed textural information is inevitably absent. In contrast, camera-based methods \cite{luo2021m3dssd, chen2017multi, huang2023tri, gao2022esgn} utilize the dense semantic information in RGB images to refine the scene representation. Due to the ambiguity of depth information in monocular images, most recent methods \cite{chen2017multi, huang2021bevdet,philion2020lift,wang2022sts} focus on multi-view images and project their features into the bird's-eye views (BEV). Subsequently, a transformer-based backbone is commonly adopted to refine the initial proposals in an end-to-end manner. However, a cross-dimensional misalignment issue still persists in camera-based methods, making it essential to design an effective multi-modal pipeline for 3D detection tasks.

\subsection{Multi-Modal 3D Detection Methods}

LiDAR points provide accurate spatial information, while cameras offer rich textural details. Therefore, this naturally motivates the integration of the two modalities for a more comprehensive and robust detection pipeline \cite{liu2023bevfusion, liang2022bevfusion, song2023graphalign}. PointPainting \cite{vora2020pointpainting} exploits the semantic segmentation information of pixels in the images to enrich the LiDAR points, which is a point-wise knowledge fusion in the early stage. Works like VirConv \cite{wu2023virtual} and EPNet \cite{huang2020epnet} attempt to incorporate image features into voxel-based or point-based features in the 3D backbone, which can be considered as middle-stage fusion. Other methods leverage multi-modal features in the refinement head for late-stage fusion \cite{chen2017multi, wu2022sparse}. MV3D \cite{chen2017multi} combines the region-wise features in 3D proposals with multi-view images, while SFD \cite{wu2022sparse} integrates the two coupled information streams via a grid-wise attentive mechanism. Nevertheless, focusing on the single stage in the detector pipeline for multi-modal fusion results in suboptimal performance because of the limitation in other stages. Therefore, we propose a multi-stage fusion strategy to enhance the representation in the whole pipeline.

\subsection{IoU-Assisted Postprocessing in the Detection}
As an essential post-processing procedure in two-stage detectors, non-maximum suppression (NMS) is commonly employed to remove duplicate bounding boxes by sorting classification scores in a descending order. However, work in \cite{jiang2018acquisition} argues that it is not appropriate to consider only the classification scores as the sorting criterion because the metric of average precision (AP) relies on the IoU with annotated bounding boxes. IoU-Uniform RCNN \cite{zhu2021iou} is a 2D detector that generates uniformly distributed proposals to train an IoU prediction branch and estimates the IoU scores in reference. 3D IoU-Net \cite{li20203d} is a pioneering work expanding IoU-assisted prediction to 3D object detection, which predicts 3D proposals from the point-wise features and corner features for subsequent 3D IoU prediction. The method in \cite{li2021voxel} further presents an IoU-guided detector by aggregating the predicted IoU with corner geometry embeddings for accurately perceiving the position of proposals. Although primary works underscore the necessity and importance of IoU-assisted confidence prediction in the voxel-based detector, research in multi-modal detectors remains relatively underexplored.

\section{Our Method}
In this section, we present our CMF-IoU detector, which performs multi-stage cross-modal 3D object detection with IoU joint prediction. As shown in Fig.~\ref{fig:framework}, the overall framework consists of three main components: (1) a bilateral cross-view enhancement backbone, designed to alleviate depth bias in pseudo points and improve distant representation in raw points; (2) an iterative voxel-point aware fine grained pooling module for the progressive refinement of proposals; and (3) an IoU joint prediction branch for more reliable bounding box confidence definition.

\subsection{Bilateral Cross-View Enhancement Backbone}
Following the previous work \cite{yin2021multimodal, wu2022casa,wu2023virtual}, we also combine the cross-modal information by combining the raw points from LiDAR with the pseudo points estimated from the images via a depth completion network. However,  the raw points are sparse but of high precision, while the estimated pseudo points are dense but usually contain noise that can not be filtered by the traditional algorithm \cite{balta2018fast}. This can result in a misalignment between the raw points and the pseudo points. Moreover, the raw points in the distinct objects are usually incomplete, which leads to sub-optimal representation and has not been addressed by previous methods \cite{shi2020pv, yin2021multimodal}. To eliminate the misalignment between the raw points and the estimated pseudo points as well as to further enhance the distant instance representation, we introduce a bilateral cross-view enhancement backbone, comprising a residual view-consistency (ResVC) and a sparse-to-distant (S2D) branch. 

\begin{figure}[ht] 
	\centering 
	\includegraphics[width=1\linewidth]{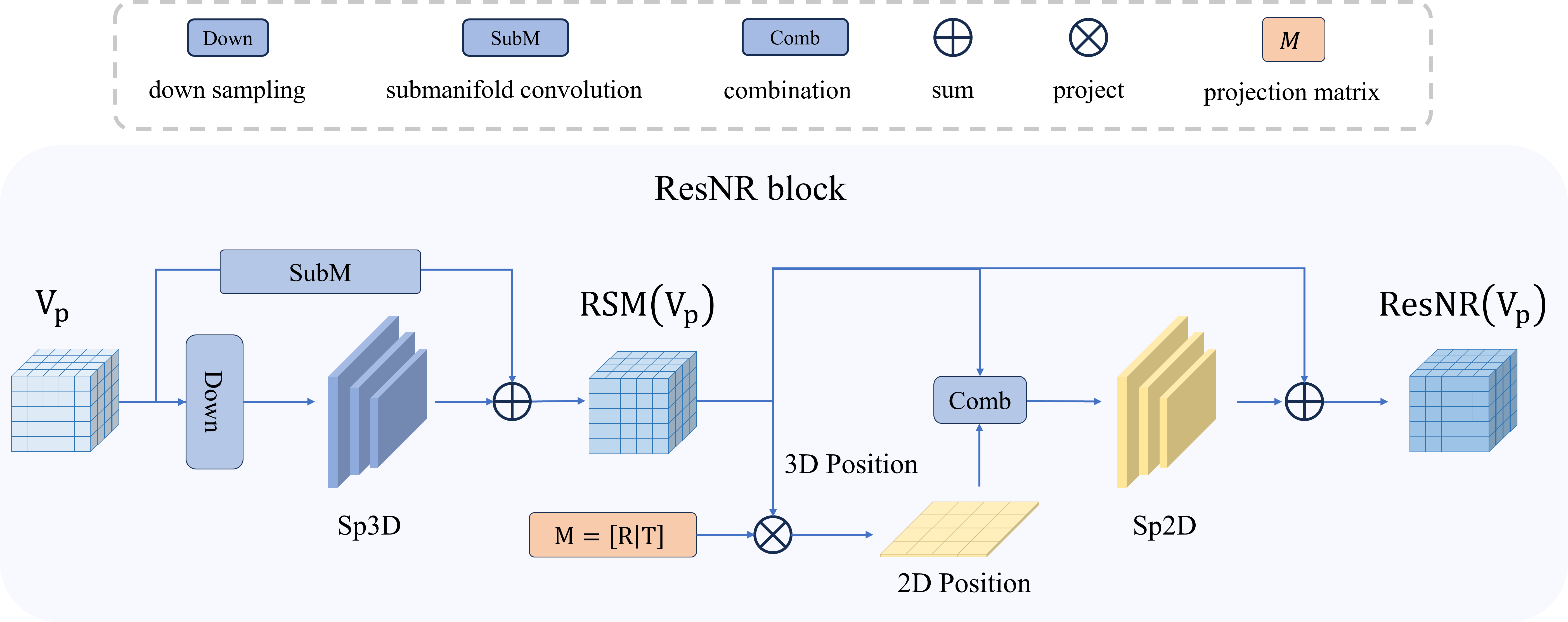} 
	\caption{Architecture of the residual noise removal (ResNR) block. The block first utilizes a residual submanifold (RSM) convolution module to encode pseudo voxels. Then, the 3D coordinates are projected into the image view and the features are subsequently encoded by 2D sparse convolutions. The output of the 2D sparse convolution is added to the output of the RSM module to form the final features encoded by the ResNR block.}
	\label{fig:resvc}  
\end{figure}

We estimate the 3D coordinates for each pixel via a depth completion network while retaining the corresponding RGB values and 2D coordinates as attributes of the pseudo points $\mathbf{P_p}$. However, noises always exist in estimated depth values and are distributed along the edges of objects in the 2D image view. Therefore, we propose the ResVC branch to mitigate the negative impact of these noisy points. Our ResVC comprises $n$ residual noise removal (ResNR) blocks, and the architecture of ResNR is shown in Fig. \ref{fig:resvc}. To be specific, we first voxelize $\mathbf{P_p}$ to get the voxel representation $\mathbf{V_p}$, and $\mathbf{V_p}$ is then encoded across different dimensions to reduce the impact of noisy points by emphasizing 2D edge features. Firstly, to facilitate gradient propagation and maintain the sparsity of features, we introduce the residual submanifold (RSM) convolution with 3D coordinates $\mathbf{P_{3d}}$, which is achieved as follows: Given $\mathbf{V_p}$, it is first proceeded with a submanifold convolution $\mathrm{SubM}(\cdot)$ with a stride of 2. At the same time, $\mathbf{V_p}$ is firstly downsampled and then proceeded with a 3D sparse convolution $\mathrm{Sp3D}(\cdot)$. The outputs of the above $\mathrm{SubM}(\cdot)$ and $\mathrm{Sp3D}(\cdot)$ are then added to form the output of the RSM, which can be formulated as:
    \begin{equation} \label{eq:eq1}
        \mathrm{RSM}(\mathbf{V_p}) = \mathrm{Sp3D}(\mathrm{Down}(\mathbf{V_p})| \mathbf{P_{3d}}) + \mathrm{SubM}(\mathbf{V_p}) .
    \end{equation}

Subsequently, the voxel coordinates $\mathbf{P_{3d}}$ in the 3D space are projected into the 2D plane by a calibration matrix $\mathbf{M}= [\mathbf{R} | \mathbf{T}]$, where $\mathbf{R}$ is the rotation matrix and $\mathbf{T}$ is the translation matrix. Then the values of the 3D features $\mathrm{RSM}(\mathbf{V_p})$ are interpolated with their corresponding 2D coordinates $\mathbf{P_{2d}} = \mathbf{M} \cdot \mathbf{P_{3d}^{T}}$ and encoded by the 2D sparse convolution $\mathrm{Sp2D}(\cdot)$ in the image view. Finally, the 3D voxel features from RSM convolution are added with the 2D convolution results to output the representation of the pseudo voxels $\mathrm{ResNR}(\mathbf{V_p})$ in the current block. Furthermore, the overall process in the ResVC branch is expressed as follows:
    \begin{equation} \label{eq:eq2}
    \begin{aligned}
        \mathrm{ResNR}(\mathbf{V_p}) &= \mathrm{Sp2D}(\mathbf{\mathrm{RSM}(\mathbf{V_p}) }| \mathbf{P_{2d}} ) + \mathbf{\mathrm{RSM}(\mathbf{V_p}) } ,\\
        \mathbf{V_p^{'}} &= \mathrm{ResVC}(\mathbf{V_p}) =\left\{ \mathrm{ResNR}\right\}_{\times n}(\mathbf{V_p}) .
    \end{aligned}
    \end{equation}

Generally, pseudo points near object boundaries (e.g., the window border of a car) often exhibit biased depth values, leading to geometric distortion when processed by conventional 3D convolutions. However, the corresponding pixel coordinates of these points in the image remain unaffected, as regions like the window edge and car body are spatially adjacent in the 2D view. By further projecting and encoding these features via 2D convolution, we can capture richer contextual cues from the image, effectively compensating for the distortion introduced during 3D encoding. As a result, the final voxel features of the window edge and car body retain regional consistency, mitigating the adverse effects of depth estimation errors.

\begin{figure}[ht] 
	\centering 
	\includegraphics[width=1\linewidth]{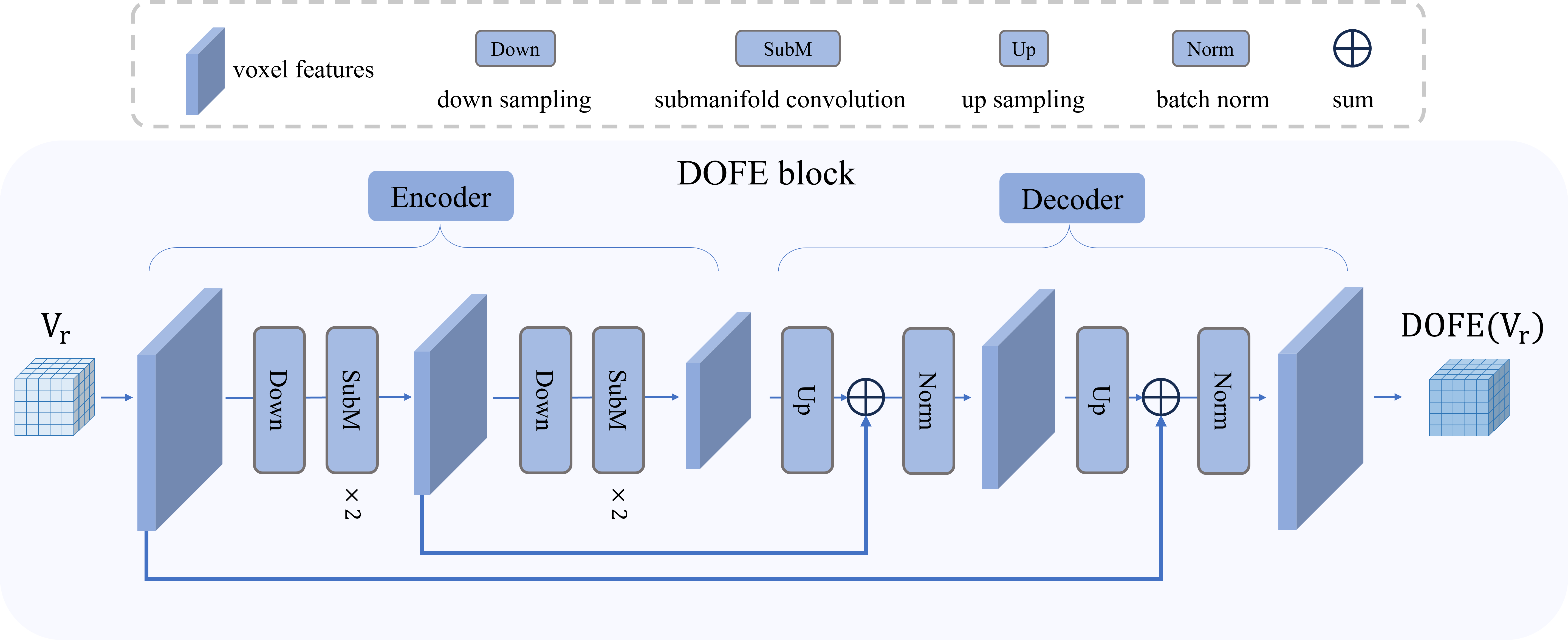}  
	\caption{Architecture of the distant object features enhancement (DOFE) block. The block contains an encoder and a decoder. The encoder utilizes two submanifold convolutions to encode each downsampled feature, then the decoder upsamples the features, and the original features are combined with skip connections for layer normalization.}
	\label{fig:dofe-layer}  
\end{figure}

When encoding the voxel information from raw points, previous works \cite{deng2021voxel, shi2020pv} typically employ sparse convolution followed by submanifold convolution in the 3D backbone to preserve the sparsity and reduce computational costs. While these approaches effectively encode information from nearby objects, they struggle to adaptively integrate features between distant foreground objects which usually suffer from sparse and incomplete points. Therefore, we design the S2D branch to enhance the representation of distant objects and preserve the sparsity in the feature space. Our S2D branch consists of $n$ distant objects features enhancement (DOFE) block, which is an encoder-decoder architecture as illustrated in Fig. \ref{fig:dofe-layer}. Given the voxelized representation $\mathbf{V_r}$ from the raw LiDAR points $\mathbf{P_r}$, the encoder layers facilitate long-range information interaction with the downsampling process, allowing to better fuse the distant objects ($>40$m) with the foreground information while maintaining the sparsity via the submanifold convolution. The decoder layers utilize the deconvolution operator to upsample the encoded features and aggregate the output with the encoder features through shortcut connection by element-wise addition. We also repeat the DOFE blocks for $n$ times to extract the final output $\mathbf{V_r^{'}}$. The total S2D branch can be formulated as:
    \begin{equation} \label{eq:eq3}
    \begin{aligned}
          \mathbf{V_r^{'}} = \mathrm{S2D}(\mathbf{V_r}) = \left\{ \mathrm{DOFE}\right\}_{\times n}(\mathbf{V_r}) .
    \end{aligned}
    \end{equation}


\subsection{Iterative Voxel-Point Aware Fine Grained Pooling}
For two-stage detectors, the refinement head is designed to further predict more accurate box locations and sizes by pooling the features in the proposals. Furthermore, we propose an iterative voxel-point aware fine grained pooling module, in which the voxel-aware part abstracts the geometric outlines while the point-aware part provides more detailed and semantic information. For more accurate prediction, we iterate this refinement stage for $T$ times. In each iteration, the pooling module can fully be aware of the raw voxel features $\mathbf{V_r^{'}}$, the pseudo voxel features $\mathbf{V_p^{'}}$, the raw points information $\mathbf{P_r}$ and the pseudo points information $\mathbf{P_p}$.

On the one hand, the voxel-aware pooling part is applied for pooling the voxel features $\mathbf{V_r^{'}}$ and $\mathbf{V_p^{'}}$ in the proposals under different grid resolutions. Taking $\mathbf{V_r^{'}}$  as an instance, in the $t$-th iteration, our method will pool the voxel features in each grid $g^t_i (i=1,...,m^3, t=1, ..., T) $, where $m$ is the grid resolution. These pooled grid features are then aggregated through the accelerated local aggregation process proposed in \cite{deng2021voxel} and the dimension will be reduced to $C$ by the multilayer perception (MLP) to obtain the final voxel pooling feature $\mathbf{g^t_r} \in \mathbb{R}^{C}$. We then concatenate all the $\mathbf{g^t_r}$ to form the feature representation for the whole scene $\mathbf{G^t_r} \in \mathbb{R}^{M \times C}$, where $M$ is the number of the proposals in the scene. Furthermore, the scene's features $\mathbf{G^t_r}$ will be passed into a cross-attention layer to enable information interaction between proposals in different refinement iteration. To be specific, the scene features $\mathbf{G^t_r}$ will be updated for $T$ times and will be loaded into a shared group $  \mathbf{S_r^{t}}$ in each iteration, denoted as $  \mathbf{S_r^{t}} = [\mathbf{S_r^{t-1}}, \mathbf{G_r^t}](t=1,\cdots, T)$, where $\mathbf{S_r^0} = \mathbf{0}$. Then, we aggregate the position embedding with $\mathbf{G_r^t}$ to generate the queries $\mathbf{Q^t_r}$ and aggregate the position embedding with shared group $\mathbf{S_r^{t-1}}$ to generate keys $\mathbf{K^t_r}$ and values $\mathbf{V^t_r}$ in the current stage. The cross-attention layer in voxel-aware pooling can be described as:
    \begin{equation} \label{eq:eq4}
    \begin{aligned}
        \mathbf{Q^t_r} &= \mathrm{FC} ( \mathbf{G_r^t} + \mathrm{PE}(\mathbf{G_r^t}) ),  \\
        \mathbf{K^t_r} , \mathbf{V^t_r} &= \mathrm{FC} ( \mathbf{S_r^{t-1}} + \mathrm{PE}(\mathbf{S_r^{t-1}}) ), \\
        \mathbf{F_{rv}^t} &= \mathrm{Softmax}(\frac{\mathbf{Q^t_r} \cdot \mathbf{K^t_r}^T}{\sqrt{D}})\mathbf{V^t_r} ,
    \end{aligned}
    \end{equation}
where $\mathrm{PE(\cdot)}$ and $\mathrm{FC(\cdot)}$ represent cosine position embedding and fully connect layers. $D$ is the channel length of $\mathbf{Q^t_r}$ and $\mathbf{K^t_r}$. $\mathbf{F_{rv}^t}$ is the final raw voxel proposal features in the $t$-th refinement iteration. Similarly, $\mathbf{F_{pv}^t}$ can be generated from the pseudo voxel features $\mathbf{V_p^{'}}$ in the same way. Then, the final voxel-aware pooling feature is calculated as $\mathbf{F_v^t}  = \mathbf{F_{rv}^{t}} \oplus \mathbf{F_{pv}^t}$, where $\oplus$ is the concatenate operator.

On the other hand, the point-aware pooling part can remarkably enhance the regression and classification accuracy in the refinement head because of the irregular but intuitive representation of 3D points. Firstly, we select the raw points $ \mathbf{P_r} \in \mathbb{R}^{N_1 \times 9}$ and pseudo points  $ \mathbf{P_p} \in \mathbb{R}^{N_2 \times 9}$ located in the predicted proposals, where $N_1$ and $N_2$ represent the points number and $9$ denotes the channel of features which contains the information of $ x,y,z $ for the 3D coordinates, $u,v$ for the corresponding 2D coordinates in the images, $r, g, b$ for RGB values and a tag position. The 2D coordinates and RGB values in $\mathbf{P_r}$ are set as 0. Then, we project $\mathbf{P_r}$ and $\mathbf{P_p}$ onto the 2D plane to enhance the local representation by aggregating their neighboring information. Specifically, the raw points $\mathbf{P_r}$ with precise $x,y$ coordinates are compressed into the BEV view $l_{bev}$, and the pseudo points $\mathbf{P_p}$ with known $u, v$ values are transformed into the camera view $l_{cam}$. It is worth noting that the locations in $l_{bev}$ and $l_{cam}$ are unbiased. We thus can effectively eliminate depth estimation errors and calibration matrix bias. Subsequently, to avoid the overlap of projection regions between $M$ different proposals, we add the proposal index $r_i, r_j \in [1,M]$ for each point, and the final 2D locations are expressed as:

\begin{equation} \label{eq:eq5}
\begin{aligned}
    l^i_{bev} &= (r_i \times h_{bev} + \mathbf{P_r}[i][0], \mathbf{P_r}[i][1]), \\
    l^j_{cam} &= (r_j \times h_{cam} + \mathbf{P_p}[j][3], \mathbf{P_p}[j][4]),
\end{aligned}
\end{equation}
where $i=1,\cdots,N_1$ and $j=1,\cdots,N_2$. $h_{bev}$ and $ h_{cam}$ represent the height in the BEV view and the camera view, respectively. By adding $r_i$ and $r_j$ in the 2D locations, Eq. \ref{eq:eq5} ensures that each patch in the projected plane only contains the points from a single proposal, and the projected pixel will inherit the feature of $\mathbf{P_r}[i]$ or $\mathbf{P_p}[j]$. The projected pixels then aggregate neighbor information by concatenating the features of all the surrounding $k \times k (k=3,5,7)$ projected pixels, and this results in a representation $f_p \in \mathbb{R}^{9 \cdot k^2}$. The local information aggregation rather than the convolution operator in 2D views can enhance each point's representation with a trade-off between effectiveness and computation.

After the local enhancement in the image view, the point feature can focus on the nearby foreground information. Then, we utilize the enhanced points features for iterative proposal refinement. In the $t$-th iteration, we sample $s$ furthest points $S^t = \{ p_1, p_2,..., p_s \}$ in each proposal to obtain the global structural representation. The sampled points will be fed into the point-aware pooling module, where the features of sampled points $f^t_{p_i} (p_i \in S^t)$ will be concatenated and encoded by a set abstraction (SA) module \cite{qi2017pointnet} to produce point-aware pooling results, which is defined as follows:
\begin{equation} \label{eq:eq6}
    \mathbf{F_p^t} = \mathrm{SA}(f^t_{p_1} \oplus \cdots \oplus f^t_{p_s} ).
\end{equation}
If there are fewer than $s$ points in the current proposal, we randomly select $s$ points with putting back. Unlike the grid division for proposals in previous works \cite{wu2022sparse, shi2020pv,deng2021voxel}, our features directly extracted by the FPS way can better focus on the overall textural representation. 

After we get the voxel-aware pooling features $\mathbf{F^t_v}$ and point-aware pooling features $\mathbf{F^t_p}$ in the $t$-th iteration, we combine them by $\mathbf{F^t} = \mathbf{F^t_v} \oplus \mathbf{F^t_p}$ to calculate the final current voxel-point aware pooling features. Then, these features will be used to predict the offset in location, size, and orientation for each 3D proposal. The updated proposals will be transferred to the $(t+1)$-th refinement iteration for more accurate target prediction as illustrated in Fig. \ref{fig:framework}(b).

\subsection{IoU Joint Branch for Confidence Prediction}
The region proposal network (RPN) in previous two-stage detectors \cite{deng2021voxel, shi2020points} is designed to generate a set of proposals that match with the foreground objects as much as possible, which is equivalent to a higher IoU value with the ground truth. However, the IoU values of proposals predicted in RPN usually obey a long-tail distribution, which results in limited performance in optimizing more accurate proposals. Meanwhile, the detection head in the refinement stage usually predicts the label scores and then filters out proposals in a decreasing order during the NMS post-processing as \cite{huang2020epnet,wu2022sparse}. However, the label scores are not exactly consistent with the actual IoU values. In this section, we design a proposals generation module and propose an IoU joint prediction branch as an auxiliary 
detection head to better preserve the proposals with high IoU and classification scores.

\begin{figure}[tbp] %%图1
	\centering  %插入的图片居中表示
	\includegraphics[width=1\linewidth]{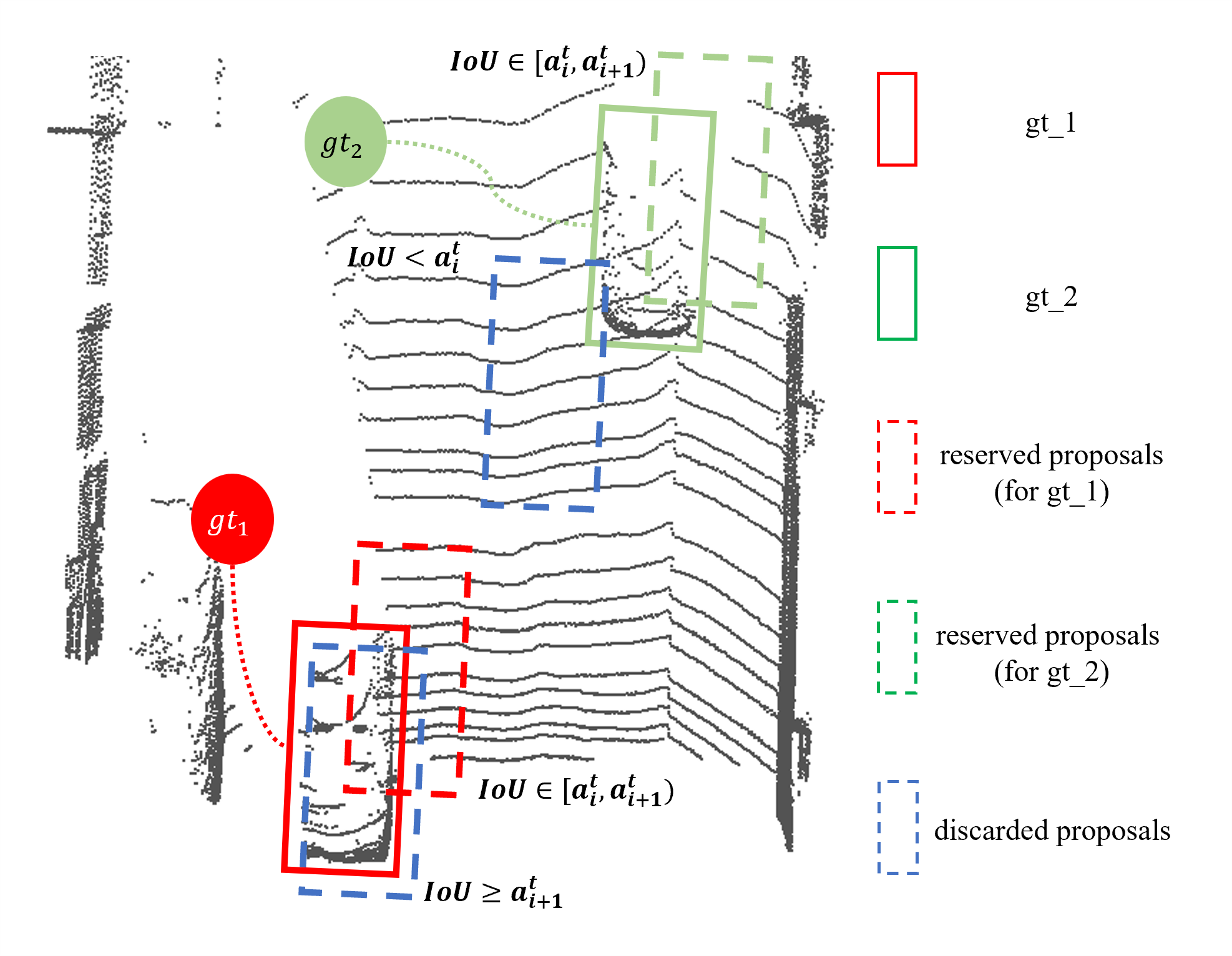}
	\caption{Illustration of uniform proposals generation with IoU values $\in [ a^t_i, a^t_{i+1} )$. In the training stage, we generate abundant proposals based on each ground-truth box, and only those whose IoU values fall within the predefined intervals are preserved.}
	\label{fig:sampling}  
\end{figure}

In the RoIs generation module, we randomly add the noise to the location, size, and yaw angle in the GT boxes to generate new boxes. Then the generated boxes are divided into four groups with incremental IoU thresholds $ [ a^t_i, a^t_{i+1} ) (i=1, \cdots,4)$. Subsequently, we only reserve random $\frac{N_g}{4}$ boxes for each group and discard the redundant boxes. Fig.\ref{fig:sampling} shows the sampling process of the proposals with the IoU threshold $ [ a^t_i, a^t_{i+1} )$. Finally, $N_g$ boxes are preserved for each ground truth box, and a total of $M \times N_g$ boxes are preserved for the current scene with the total $M$ proposals. During the iteration in the refinement head, the sampled IoU thresholds are also increased gradually to motivate the network to better optimize the proposals with high IoU. Notably, the GT-based proposal generation technique is only used in training.

Since classification scores alone may not accurately reflect the actual IoU values, it is essential to introduce a dedicated branch for IoU prediction. As shown in Fig.~\ref{fig:framework}, the regression, classification, and IoU values are calculated in each iteration of the refinement head, while the final IoU prediction is performed only for the last refined proposals. The resulting IoU score is then combined with the classification score to form the final confidence score. Notably, our GT-based proposal generation strategy provides a large number of proposals with uniformly distributed IoU values, which significantly enhances the generalization ability of the IoU prediction branch across varying levels of localization accuracy.

\subsection{Loss Function}
Compared with the original detectors, we innovatively propose the iterative voxel-point aware fine grained pooling and IoU joint prediction in the refinement stage. The loss functions in all the iterations are accelerated to acquire the total loss function in our refinement stage, which is formulated as:
\begin{equation} \label{eq:eq12}
    L_{Total} = \sum_{t=1}^{T} L(\mathcal{M}_t,\hat{B}),   
\end{equation}
where $\hat{B}$ represents the annotated boxes, and $\mathcal{M}_t$ denotes the model output in the $t$-th refinement iteration. In each iteration, the loss function $L(\mathcal{M}_t,\hat{B})$ is composed of the comprehensive loss $L^t_c$ , the voxel-aware loss $L^t_v$ , and the pool-aware loss $L^t_p$ which are calculated from the proposals predicted by $\mathbf{F^t}$, $\mathbf{F^t_v}$ and $\mathbf{F^t_p}$, respectively. The loss function in each refinement iteration can be denoted as:
\begin{equation}
    L(\mathcal{M}_t,\hat{B}) = (L^t_c + w_1 L^t_v + w_2 L^t_p).
\end{equation}
where $w_1=w_2=0.5$ are the predefined weights. Furthermore, all of the $L^t_c$, $L^t_v$, and $L^t_p$ contain the smooth-L1 loss for both the regression scores and the IoU scores, and the binary cross-entropy (BCE) loss for classification scores. Taking $L^t_c$ as an instance, the function is illustrated as:
\begin{equation}
    L^t_c = L_{SL1}(I^t_c ) + L_{SL1}(R^t_c) + L_{BCE}(C^t_c),
    \label{eq:eq11}
\end{equation}
 where $I^t_c$, $R^t_c$ and $C^t_c$ are the IoU, regression and classification values predicted by $\mathbf{F^t}$. The formula of $L^t_p$ and $L^t_v$ is same as $L^t_c$ with the prediction results of $\mathbf{F^t_p}$ and $\mathbf{F^t_v}$, respectively.

\section{Experiments}

\subsection{Datasets and Evaluation Metrics}

\subsubsection{KITTI dataset} 

The KITTI dataset \cite{geiger2012we} provides LiDAR point clouds and only front-view camera images. The dataset consists of 7,481 paired training samples and 7,518 paired test samples. Following the previous works \cite{deng2021voxel,shi2020points,shi2020pv}, we divide the original training images into a training set and a validation set. There are 3,712 paired images in the training set and 3,769 paired images in the validation set. The standard evaluation metrics designed for object detection are 3D and BEV average precision (AP) under 40 recall thresholds (R40). In this work, we evaluate our method on the three major categories with IoU thresholds of 0.7 for car, 0.5 for pedestrian and 0.5 for cyclist. To compare our results with other state-of-the-art methods on the KITTI 3D detection benchmark, we train our method on all 7,481 images for testing. 

\subsubsection{nuScenes dataset}

The nuScenes dataset \cite{caesar2020nuscenes} is a large-scale 3D detection benchmark that encompasses more complex scenarios, such as nighttime and various urban environments, compared with the KITTI dataset. The surrounding information is captured with six multi-view cameras and a 32-channel LiDAR sensor, providing 360-degree object annotations for 10 object classes. It consists of 700 training scenes, 150 validation scenes and 150 testing scenes. To evaluate detection performance, the primary metrics mean average precision (mAP) and the nuScenes detection score (NDS) are employed. These metrics assess a method's detection accuracy in terms of classification, bounding box location, size, orientation, attribute, and velocity.

\subsubsection{Waymo dataset}

The Waymo dataset \cite{sun2020scalability} includes 230k annotated samples, which are partitioned into 160k training samples, 40k validation samples and 30k samples for testing. The whole scene in the point clouds covers a large reception range of $150 m \times 150 m$. The main evaluation metric is the average precision (AP). The detection difficulty of each object is divided into two categories: Level 1 (L1) for objects containing more than five points and Level 2 (L2) for those containing at least one point.

\subsection{Implementation Details}

Our CMF-IoU is meticulously trained from scratch using the Adam optimizer with the one-cycle learning rate policy. The optimizer utilizes a weight decay of 0.01 and a momentum value of 0.9. We adopt the depth completion network PENet \cite{hu2021penet} for the KITTI dataset and MVP \cite{yin2021multimodal} for nuScenes and Waymo datasets to generate the pseudo points. To enable efficient training on the datasets, we utilize two NVIDIA RTX 4090D GPUs. Due to the distinct evaluation metrics and characteristics in the above datasets, we employ different implementations for each dataset. For the KITTI dataset, we replace the backbone and refinement module in the VoxelRCNN \cite{deng2021voxel} with our proposed methods, which requires approximately 5 hours for training 40 epochs with a batch size of 2 and a learning rate of 5e-3. The entire detection scale is set within the range \([-40, 40] \times [0, 70.4] \times [-3, 1]\) meters in 3D space. We also adopt the same synchronized data augmentation solution as SFD \cite{wu2022sparse}. For the nuScenes dataset, our model utilizes the baseline of VoxelNeXt \cite{chen2023voxelnext} and the dense head in SAFDNet \cite{zhang2024cvpr} and replaces the backbone with ours, which requires approximately 50 hours for training 20 epochs with a learning rate of 1e-3. The detection range is set to \([-54, 54] \times [-54,54] \times [-5,3]\) meters. We adopt the same data augmentation configuration in the training stage as that in previous works \cite{chen2023voxelnext, yin2021center}.  For the Waymo dataset,  we apply our method on the PV-RCNN \cite{shi2020pv}, and the detection scale is set to \([-75.2, 75.2] \times [-75.2, 75.2] \times [-2,4]\) meters. We train on the 20\% datasets with a learning rate of 1e-2 and test on the full validation dataset. In the refinement stage, we set the iteration number $T=3$ for a trade-off between computation and efficiency. Furthermore, in the $t$-th iteration, the predefined IoU scale $[a^t, 1)$ is split into four uniform ranges, and the lower bound $a^t$ is increased during the iteration. In the inference stage, we employ our IoU-classification balanced metric for the NMS post-processing to suppress redundant proposals. These configurations are determined empirically to ensure stable convergence and optimal performance on the respective dataset.

\subsection{Comparison with the State-of-the-Art Methods}

\subsubsection{Evaluation on the KITTI test dataset}
We compare our CMF-IoU against the state-of-the-art methods in 3D and BEV AP metrics on the KITTI test dataset. As shown in Table \ref{tab_KITTI_test}, our method achieves superior performance, which produces remarkably competitive results at three difficulty levels, with 91.92\%, 85.14\%, 80.63\% in 3D APs and 95.72\%, 92.07\%, 87.25\% in BEV APs, respectively. Our CMF-IoU achieves better performance than the baseline VoxelRCNN\cite{deng2021voxel}, with improvements of 1.02\%, 3.52\%, 3.57\% in 3D APs and 0.87\%, 3.24\%, 1.12\% in  BEV APs across the three difficulty levels. Compared with the multi-modal fusion method EPNet \cite{huang2020epnet}, CMF-IoU achieves better performance with improvements of 2.11\%, 5.86\% and 6.04\% in 3D APs. CMF-IoU performs particularly well on the easy and moderate levels, which encompass most autonomous driving scenarios.

%表1  KITTI上的测试集比较
\begin{table*}[t]
\scriptsize
\large
\centering
\caption{ Comparison on the KITTI test set. The results are reported with the mAP with 0.7 IoU threshold in car, 0.5 IoU threshold in pedestrian and cyclist with 40 recall threshold. ‘L’ and ‘C’ represent LiDAR and camera, respectively.}
\renewcommand\arraystretch{1}
\resizebox{\textwidth}{!}{
\begin{tabular}{@{}@{\extracolsep{\fill}}!{\color{white}\vline}l|c|c|c|c|c|c|c|c|c|c|c|c|c|c|c|c|c|c|c @{}}
\toprule
\multirow{3}{*}{Method} & \multirow{3}{*}{Modality} & \multicolumn{6}{c|}{Car}                                 & \multicolumn{6}{c|}{Pedestrain}                & \multicolumn{6}{c}{Cyclist}                  \\ \cmidrule(l){3-20} 
    &                           & \multicolumn{3}{c|}{AP$_{3D}$ (\%)} & \multicolumn{3}{c|}{AP$_{BEV}$ (\%)} & \multicolumn{3}{c|}{AP$_{3D}$ (\%)} & \multicolumn{3}{c|}{AP$_{BEV}$ (\%)} & \multicolumn{3}{c|}{AP$_{3D}$ (\%)} & \multicolumn{3}{c}{AP$_{BEV}$ (\%)}\\ \cmidrule(l){3-20}
   & & Easy & Mod & Hard & Easy & Mod & Hard & Easy & Mod & Hard & Easy & Mod & Hard & Easy & Mod & Hard & Easy & Mod & Hard     \\ \midrule
   DSGN \cite{chen2020dsgn} & C & 73.50 & 52.18 & 45.14 & 82.90 & 65.05 & 56.60 & - & - & - & - & - & - & - & - & - & - & - & -\\
   YOLOStereo3D \cite{yolostereo3d} & C & 65.68 & 41.25  & 30.42 & - & - & - &  28.49 & 19.75 & 16.48 & - & - & - & - & - & - & - & - & -  \\
   ESGN \cite{gao2022esgn} & C & 46.39 & 65.80 & 38.42 & - & - & - & - & - & - & - & - & - & - & - & - & - & - & - \\
   PV-RCNN\cite{shi2020pv} & L & 90.25 & 81.43 & 76.82 & 94.98 & 90.65 & 86.14 & 52.17 & 43.29 & 40.29 & 59.86 & 50.57 & 46.74 & 78.60 & 63.71 & 57.65 & 82.49 & 68.89 & 62.41 \\
   VoxelRCNN\cite{deng2021voxel} & L & 90.90 & 81.62 & 77.06 & 94.85 & 88.83 & 86.13 & - & - & - & - & - & - & - & - & - & - & - & -        \\
   LGSLNet \cite{qiao2024local} & L & 90.51 & 82.16 & 79.33 & 94.35 & 90.85 & 88.27 & - & - & - & - & - & - & - & - & - & - & - & - \\
   % SECOND\cite{yan2018second} & L & 83.13 & 73.66 & 66.20 & 88.07 & 79.37 & 77.95 & 51.07 & 42.56 & 37.29 & 55.10 & 46.27 & 44.76 &70.51 & 53.85 & 46.90 & 73.67 & 56.04 & 48.78 \\
   % SE-SSD\cite{zheng2021se} & L & 91.49 &	82.54 &	77.15 & 95.68 &	91.84 &	86.72 & - & - & - & - & - & - & - & - & - & - & - & -\\
   PointRCNN\cite{shi2019pointrcnn} & L &  85.94 &	75.76 &	68.32 & - &	- &	- & 49.43 &	41.78 &	38.63 & - &	- &	- & 73.93 &	59.60 &	53.59 & - &	- &	- \\
   PointPillar\cite{lang2019pointpillars} & L &  79.05 & 74.99 & 68.30 & 88.35 & 86.10 & 79.83 & 52.08 & 43.53 & 41.49 & 58.66 & 50.23 & 47.19 & 75.78 & 59.07 & 52.92 & 79.14 & 62.25 & 56.00 \\
   Part-A2 \cite{shi2020points} & L & 85.98 & 77.86 & 72.00 & 89.52 & 84.76 &  81.47 & - & - & - & - & - & - & 78.58 & 62.73 & 57.74 &  81.91 &  68.12 & 61.92 \\
   3D IoU-Net\cite{li20203d} & L & 87.96 &	79.03 &	72.78 & 94.76 &	88.38 &	81.93 & - & - & - & - & - & - & - & - & - & - & - & -        \\ \midrule
    MV3D\cite{chen2017multi} & L\&C &  74.97 & 63.63 &	54.00 & 86.62 &	78.93 &	69.80 & - & - & - & - & - & - & - & - & - & - & - & - \\
   EPNet\cite{huang2020epnet} & L\&C & 89.81 & 79.28 & 74.59 & 94.22 & 88.47 & 83.69 & - & - & - & - & - & - & - & - & - & - & - & -        \\
    % F-PointNet\cite{qi2018frustum}& L\&C & 81.20 & 70.39 & 62.19 & 88.70 & 84.00 & 75.33 & 51.21 & 44.89 & 40.23 & 58.09 & 50.22 & 47.20 & 71.96 & 56.77 & 50.39 & 75.38 & 61.96 & 54.68        \\
    % Focals Conv\cite{chen2022focal}& L\&C & 90.55 & 82.28 & 77.59 & 92.67 & 89.00 & 86.33 & - & - & - & - & - & - & - & - & - & - & - & -        \\
    VoPiFNet\cite{wang2024vopifnet} & L\&C & 88.51 & 80.97 & 76.74 & 93.94 &	90.52 &	86.25 & \textbf{54.65} & 48.36 & 44.98 & \textbf{60.07} &	52.41 &	50.28 & 77.64 & 64.10 & 58.00 & 80.83 &	67.66 &	61.36    \\
    % Graph-VoI\cite{yang2022graph} & L\&C & 91.89 & 83.27 & 77.78 & 95.69 & 90.10 & 86.85 & - & - & - & - & - & - & - & - & - & - & - & -    \\
    GraphAlign++ \cite{song2023graphalign++} & L\&C & 90.98 & 83.76 & 80.16 & 93.97 & 91.83 & 88.65 & 53.31 & 47.51 & 42.26 & 57.71 & 51.35 & 46.33 & 79.58 & 65.21 & 55.68 & 82.68 & 68.56 & 63.33 \\
    % 3D-CVF\cite{yoo20203d} & L\&C & 89.20 &	80.05 &	73.11 & 93.52 &	89.56 &	82.45 & - & - & - & - & - & - & - & - & - & - & - & - \\
    SFD\cite{wu2022sparse} & L\&C & 91.73 & 84.76 & 77.92 & 95.64 & 91.85 & 86.83 & - & - & - & - & - & - & - & - & - & - & - & -  \\
    LoGoNet\cite{li2023logonet} & L\&C & 91.80 & 85.06 & \textbf{80.74} & 95.48 & 91.52 & 87.09 & 53.07 & 47.43 & 45.22 & 58.24 &	52.06 &	49.87 & 84.47 & 71.70 & \textbf{64.67} & 85.85 &	\textbf{74.92} & \textbf{67.62}  \\ 
    VirConv-L\cite{wu2023virtual}& L\&C & 91.41 & 85.05 & 80.22 & 95.53 & 91.95 & 87.07 & - & - & - & - & - & - & - & - & - & - & - & - \\ \midrule
    % VoxelRCNN* & L & 90.76 & 81.69 & 77.42 & 92.89 & 89.97 & 84.69 & 52.57 & 44.86 & 39.09 & 57.66 & 49.32 & 44.15 & 77.54 & 64.00 & 53.15 & 79.68 & 67.56 & 62.70        \\
    \rowcolor{blue!10} CMF-IoU (Ours)& L\&C  & \textbf{91.92} & \textbf{85.14} & 80.63 & \textbf{95.72} & \textbf{92.07} & \textbf{87.25} & 53.26 & \textbf{49.23} & \textbf{46.19} & 58.62 & \textbf{52.48} &	\textbf{50.31} & \textbf{85.56} & \textbf{71.84} & 63.43 & \textbf{87.21} &	72.87 &	66.85 \\
     \bottomrule
\end{tabular}
}
\label{tab_KITTI_test}
\end{table*}

\subsubsection{Evaluation on the KITTI validation dataset}
Table \ref{tab_KITTI_val} presents the quantitative comparison with the state-of-the-art methods in 3D and BEV AP metrics of car on the KITTI validation dataset. It is clear that our CMF-IoU also outperforms both single-modal and multi-modal methods. In detail, our method achieves better or comparable performance at all difficulty levels of 3D and BEV detection, including 95.52\%, 89.01\%, 87.21\% in 3D APs and 95.84\%, 93.04\%, 91.16\% in BEV APs. Compared with the model VoxelRCNN \cite{deng2021voxel}, our method obtains improvements of 3.14\%, 3.72\% and 4.35\% across the three levels in 3D APs, respectively. Compared with the multi-modal methods, our CMF-IoU achieves impressive detection accuracy in the hard difficulty levels (containing fewer points), which demonstrates that our method can be adapted well to long-range objects.

%表2  KITTI上的验证集比较
\begin{table}[t]
\scriptsize
\centering
\caption{Comparison on the KITTI validation dataset for the car class. ‘L’ and ‘C’ represent LiDAR and the camera, respectively.}
\renewcommand\arraystretch{1}
\setlength{\tabcolsep}{1.3mm}{

\begin{tabular}{@{}@{\extracolsep{\fill}}!{\color{white}\vline}l|c|c|c|c|c|c|c @{}}
\toprule
\multirow{2}{*}{Method}& \multirow{2}{*}{Modality}& \multicolumn{3}{c|}{AP$_{3D} (\%)$}   & \multicolumn{3}{c}{AP$_{BEV} (\%)$}                                        \\ \cmidrule(r){3-8} 
&    & Easy   & Mod.  & Hard & Easy & Mod. & Hard  \\ \midrule
DSGN \cite{chen2020dsgn} & C &  72.31 & 54.27 & 47.71 & 83.24 & 63.91 & 57.83 \\
ESGN \cite{gao2022esgn} & C & 52.33 & 72.44 & 43.74 & 63.86 & 82.29 & 54.63 \\
% SECOND \cite{yan2018second} & L  & 87.43 & 76.48  & 69.10 & \_  & \_  & \_  \\ 
PointRCNN \cite{shi2019pointrcnn}  & L  & 88.88 & 78.63 & 77.38 & \_  & \_  & \_ \\
3D IoU-Net  \cite{li20203d} & L & 89.31 & 79.26 & 78.68 & - & - & - \\
VoxelRCNN \cite{deng2021voxel}& L & 92.38 & 85.29 & 82.86 & 95.52 & 91.25 & 88.99\\
% Part-A2 \cite{shi2020points} & L & 89.47 & 79.47 & 78.54 & 90.42 & 88.61 & 87.31 \\
PV-RCNN \cite{shi2020pv}& L & 92.57 & 84.83  & 82.69 & 95.76 & 91.11 & 88.93 \\
LGSLNet \cite{qiao2024local} & L & 93.01 & 85.94 & 85.59 & - & - & - \\ \midrule
% CT3D \cite{sheng2021improving}  & L  & 92.85& 85.82 & 83.46 & 96.14	& 91.88	 & 89.63 \\ \midrule
MV3D \cite{chen2017multi} & L\&C & 71.29 & 62.68 & 56.56 & \_ & \_ & \_ \\
EPNet \cite{huang2020epnet} & L\&C & 92.28 & 82.59 & 80.14 & 95.51 & 91.47 & 91.16 \\
LoGoNet\cite{li2023logonet} & L\&C & 91.80 & 85.06 & 80.74 & \_ & \_  & \_  \\
GraphAlign++ \cite{song2023graphalign++} & L\&C & 92.58 & 87.01 & 84.68 & 95.65 & 92.82 & \textbf{91.41} \\
VirConv-L \cite{wu2023virtual}& L\&C & 93.36 & 88.71 & 85.83 & \_ & \_  & \_  \\
TED-M \cite{wu2023transformation} & L\&C & 95.25 & 88.94 & 86.73 & \_ & \_  & \_  \\
\midrule        
\rowcolor{blue!10} CMF-IoU (Ours)& L\&C &  \textbf{95.52} & \textbf{89.01} & \textbf{87.21} & \textbf{95.84} & \textbf{93.04} & 91.16  \\ 
\bottomrule
\end{tabular}
}
\label{tab_KITTI_val}
\end{table}

\subsubsection{Evaluation on the nuScenes test dataset}
As shown in Table \ref{tab_nuScens_test}, we conduct experiments on the more challenging nuScenes test dataset to further validate the effectiveness of our CMF-IoU. Specifically, CMF-IoU achieves state-of-the-art performance with a mAP of 69.8\% and NDS of 72.6\%. Furthermore, we emphasize the performance improvements in categories such as ``C.V.", ``Barrier'', ``Ped.", and ``T.C.", where CMF-IoU achieves significant gains over previous methods. In particular, CMF-IoU shows improvements of 4.6\% and 9.3\% over VoxelNeXt in the ``Ped." and ``T.C.", respectively. These results validate the ability of CMF-IoU in detecting 3D instances, especially for categories that are typically characterized as small objects.

%表3  NuScenes上的测试集比较
\begin{table*}[t]
\centering
\caption{Comparison on the nuScenes test dataset. 'C.V.', 'Motor.', 'Ped.', and 'T.C.' represent construction vehicle, motorcycle, pedestrian, and traffic cone, respectively.}
\renewcommand\arraystretch{1}  % 调整行间距
\resizebox{0.8\textwidth}{!}{  % 保持宽度不变
\begin{tabular}{c|c|c|c|c|c|c|c|c|c|c|c|c}
\toprule
Method & mAP  & NDS  & Car  & Truck & C.V. & Bus  & Trailer & Barrier & Motor. & Bicycle & Ped. & T.C. \\ 
\midrule
PointPillars\cite{lang2019pointpillars} & 30.5& 45.3 &68.4 &23.0& 4.1 &28.2 &23.4 &38.9 &27.4& 1.1& 59.7& 30.8 \\
CenterPoint \cite{yin2021center} & 58.0 & 65.5 & 84.6 & 51.0 & 17.5 & 60.2 & 53.2 & 70.9 & 53.7 & 28.7 & 83.4 & 76.7 \\
PointPainting \cite{vora2020pointpainting} & 46.4 & 58.1 & 77.9 & 35.8  & 15.8 & 36.2 & 37.3 & 60.2 & 41.5 & 24.1 & 73.3 & 62.4 \\
VoxelNeXt \cite{chen2023voxelnext} & 64.5 & 70.0 & 84.6 & 53.0 & 28.7 & 64.7 & 55.8 & 74.6 & 73.2 & 45.7 & 85.8 & 79.0 \\
S2M2-SSD  \cite{zheng2022boosting} & 62.9 & 69.3 & 86.3 & 56.0  & 26.2 & 65.4 & 59.8    & 75.1    & 61.6   & 36.4  & 84.6 & 77.7 \\
AFDetV2  \cite{hu2022afdetv2} & 62.4 & 68.5 & 86.3 & 54.2  & 26.7 & 62.5 & 58.9    & 71.0    & 63.8   & 34.3  & 85.8 & 80.1 \\
MVP  \cite{yin2021multimodal} & 66.4 & 70.5 & 86.8 & 58.5  & 26.1 & 67.4 & 57.3  & 74.8    & 70.0   & 49.3 & 89.1 & 85.0 \\
VISTA  \cite{deng2022vista} & 63.0 & 69.8 & 84.4 & 55.1  & 25.1 & 63.7 & 54.2    & 71.4    & 70.0   & 45.4  & 82.8 & 78.5 \\
AutoAlign\cite{chen2022autoalign} & 65.8 & 70.9 & 85.9 & 55.3 & 29.6 & 67.7 & 55.6 & - & 71.5 & 51.5 & 86.4 & - \\
Focals Conv-F \cite{chen2022focal} & 67.8 & 71.8 & 86.5 & 57.5 & 31.2 & 68.7 & 60.6 & 72.3 & 76.4 & 52.5 & 87.3 & 84.6 \\
TransFusion \cite{bai2022transfusion} & 68.9 & 71.7 & 87.1 & 60.0 & 33.1 & 68.3 & 60.8 & 78.1 & 73.6 & 52.9 & 88.4 & 86.7 \\
GraphAlign++ \cite{song2023graphalign++} & 68.5 & 72.2 &  87.5 & 58.5 & 32.3 & 68.9 & 58.3 & 74.3 & 76.4 & 53.9 & 88.3 & 86.3 \\
FocalFormer3D \cite{focalformer3d} & 68.7 & \textbf{72.6} & 87.2 & 57.0 & 34.4 & 69.6 & 64.9 & 77.8 & 76.2 & 49.6 & 88.2 & 82.3 \\
% BEVFusion(MIT) \cite{liu2023bevfusion} & 70.2 & 72.9 & 88.6 & 60.1 & 39.3 & 69.8 & 63.8 & 80.0 & 74.1 & 51.0 & 89.2 & 86.5\\
BEVFusion \cite{liang2022bevfusion} & 69.2 & 71.8 & 88.1 & 60.9 & 34.4 & 69.3 & 62.1 & 78.2 & 72.2 & 52.2 & 89.2 & 85.2 \\
% MSMDFusion \cite{Jiao_2023_CVPR} & 71.5 & 74.0 & 88.4 & 61.0 & 35.2 & 71.4 & 64.2 & 80.7 & 76.9 & 58.3 & 90.6 & 88.1 \\ 
% CMT \cite{yan2023cross} & 70.4 & 73.0 & 87.2 & 61.5 & 37.5 & 72.4 & 62.8 & 74.7 & 79.4 & 58.3 & 86.9 & 83.2 \\
\midrule
\rowcolor{blue!10} CMF-IoU (Ours)& \textbf{69.8} & \textbf{72.6} & 87.5 & 59.0 & 36.2 & 69.5 & 65.1 & 78.9 & 74.9 & 48.2 & 90.4 & 88.3 \\ 
\bottomrule
\end{tabular}
}
\label{tab_nuScens_test}
\end{table*}

%表4  NuScenes上的测试集比较
\begin{table*}[!ht]
\centering
\caption{Comparison on the nuScenes validation dataset. `C.V.', `Ped.', `Motor.', and `T.C.' represent construction vehicle, pedestrian, motorcycle, and traffic cone, respectively. For a fair comparison, none of the methods utilizes test-time augmentation.}
\renewcommand\arraystretch{1}  % 调整行间距
\resizebox{0.83\textwidth}{!}{  % 保持宽度不变
\begin{tabular}{l|c|c|c|c|c|c|c|c|c|c|c|c}
\toprule
Method & NDS  & mAP  & Car  & Truck & Bus  & Trailer  & C.V. & Ped. & Motor. & Bicycle& T.C. & Barrier  \\ 
\midrule
CenterPoint\cite{yin2021center} & 66.5 & 59.2 & 84.9 & 57.4 & 70.7 & 38.1 & 16.9 & 85.1 & 59.0 & 42.0 & 69.8 & 68.3\\
VoxelNeXt \cite{chen2023voxelnext} & 66.7 & 60.3 & 83.7 & 56.0 & 68.5 & 38.1 & 21.1 & 85.2 & 63.8 & 48.9 & 69.6 & 69.0 \\
TransFusion \cite{bai2022transfusion} & 70.1 & 65.5 & 86.9 & 60.8  & 73.1 & 43.4 & 25.2 & 87.5 & 72.9 & 57.3 & 77.2 & 70.3 \\
DSVT \cite{wang2023dsvt}   & 71.1 & 66.4 & 87.4 & 62.6 & 75.9 & 42.1 & 25.3 & 88.2 & 74.8 & 58.7 & 77.9 & 71.0\\
HEDNet  \cite{zhang2023hednet} & 71.4 & 66.7 & 87.7 & 60.6 & 77.8 & 50.7 &  28.9 & 87.1 & 74.3  & 56.8 & 76.3 & 66.9 \\
FUTR3D \cite{chen2022futr3d} & 68.0 & 64.2 & 86.3 & 61.5 & 71.9 & 42.1 & 26.0 & 82.6 & 73.6 & 63.3 & 70.1 & 64.4 \\
TranFusion \cite{bai2022transfusion} & 71.3 & 67.5 & -& -& -& -&- & -& -& -& -&- \\
BEVFusion-MIT \cite{liu2023bevfusion} & 71.4 & 68.5 & -& -& -& -&- & -& -& -& -&- \\
BEVFusion-PKU \cite{liang2022bevfusion} & 71.0 & 67.9 & 88.6 & 65.0 & 75.4 & 41.4 & 28.1 & 88.7 & 76.7 & 65.8 & 76.9 & 72.2 \\ \midrule
% MSMDFusion \cite{Jiao_2023_CVPR} & 72.1 & 69.3 & -&- & -& -&- &- &- &- & -& -\\ \midrule
% CMT \cite{yan2023cross} & 72.9 & 70.3 & -&- & -& -& -& -& -&- &- &- \\ \midrule
% UniTR \cite{wang2023unitr} & 73.3 & 70.5 & & & & & & & & & & \\ \midrule
\rowcolor{blue!10} CMF-IoU (Ours)& \textbf{72.4} & \textbf{69.1} & 88.2 & 62.4 & 77.5 & 48.5 & 28.9 & 90.1 & 77.2 & 64.7 & 82.1 & 71.3 \\
\bottomrule
\end{tabular}
}
\label{tab_nuscenes_valid}
\end{table*}

\subsubsection{Evaluation on the nuScenes validation dataset}
To demonstrate the applicability of our CMF-IoU framework, we also conduct the comparative experiment on the nuScenes validation dataset. As shown in Table \ref{tab_nuscenes_valid}, our method outperforms the state-of-the-art method BEVFusion-MIT \cite{liu2023bevfusion} by 0.6\%, 1.0\% in mAP and NDS metrics, respectively. Additionally, our method shows significant improvements on the ``Ped.'' and ``T.C.'' categories, which are relatively smaller than others.

\subsection{Ablation Study}
In this subsection, we present additional ablation experiments on the KITTI, nuScenes, and Waymo validation datasets. We first demonstrate the effectiveness of the proposed method on different datasets. Then, we introduce point noises and calibration errors into the scenes to evaluate the robustness and reliability of our approach. Additionally, we investigate the impact of each designed module by reporting the results under different module combinations and parameter settings.

\subsubsection{Effectiveness of our proposed modules}
To demonstrate the generalization and effectiveness of our approach, we conduct ablation studies under different baselines, datasets, and component combinations. As shown in Table \ref{tab_baseline_comp}, our method consistently boosts the performance across various baselines. Notably, it achieves substantial gains over VoxelNeXt \cite{chen2023voxelnext} and CenterPoint \cite{yin2021center}, with improvements of approximately 5\% in NDS and 3\% in mAP. Compared to the strong baseline SAFDNet \cite{zhang2024cvpr}, our method still delivers a 2.1\% gain in NDS. Table \ref{tab_component_comp_kitti} further demonstrates the generalization of each module in the KITTI and Waymo datasets. Furthermore, Table \ref{tab_pre_component_comp} presents the comparisons with previous strategies for multi-modal fusion or IoU joint prediction, which also validates the superior performance of our method.

\subsubsection{Comparison of different branch combinations in the 3D backbone}
To evaluate the superiority of our bilateral cross-view enhancement backbone, we provide fair comparison experiments using different branch combinations. As shown in Table \ref{3b_backbone_ablation}, the initial 3D APs of the basic model for the easy, moderate and hard categories are only 94.78\%, 88.45\% and 85.99\%, respectively. The consecutive addition of the S2D branch and the ResVC branch can both improve the performance. Furthermore, the 3D APs in the hard difficulty level can achieve 88.14\% with only the addition of the ResVC branch. When both branches are incorporated, the final AP metrics in three difficulty levels can achieve the amplification of 1.23\%, 0.52\% and 2.32\%, respectively. The results demonstrate that the S2D branch can facilitate remote information integration. Besides, the misalignment between the pseudo points and raw points, especially located in the hard instances, can be effectively eliminated by our proposed ResVC branch. 

\begin{table}[!t]
\centering
\caption{The improvement of our method when integrated with different baselines on the nuScenes validation dataset. `*' indicates the reproduced results using the dynamic voxelization setting.}
\renewcommand\arraystretch{1}
\resizebox{0.95\linewidth}{!}{
\centering
\begin{tabular}{c|c|c|c|c|c}
\toprule
 Method & mAP  & NDS & Car & Ped. & Bicycle \\ 
\midrule
 VoxelNeXt \cite{chen2023voxelnext} & 60.3 & 66.7 & 83.7& 85.2& 48.9\\
 CMF-IOU & 65.8 \textit{(+5.5)} & 70.8 \textit{(+4.1)} & 87.3& 88.4&  56.8\\
 \midrule
CenterPoint* \cite{yin2021center} & 61.8 & 68.5 & 86.0 & 86.3& 52.0\\
CMF-IOU & 66.7 \textit{(+4.9)} & 71.2 \textit{(+2.7)} & 86.9 & 88.2 & 59.5\\
\midrule
SAFDNet \cite{zhang2024cvpr} & 67.0 & 71.1 & 87.5& 87.6& 59.7\\
CMF-IOU & 69.1 \textit{(+2.1)}& 72.4 \textit{(+1.3)} & 88.2& 90.1& 64.7\\
\bottomrule
\end{tabular}
}
\label{tab_baseline_comp}
\end{table}

\begin{table}[!t]
\centering
\caption{Effectiveness of the modules in our pipeline. For fair comparison, we utilize the same input points and vary the component combinations. ``Init'' indicates the baseline models, where VoxelRCNN is used for KITTI and PV-RCNN is adopted for Waymo. ``MS Fusion'' denotes the multi-stage fusion strategy. ``Veh.'', ``Ped.'' and ``Cyc.'' indicate vehicle, pedestrian and cyclist, respectively.}
\renewcommand\arraystretch{1}
\resizebox{1.0\linewidth}{!}{
\large
\centering
\begin{tabular}{l|c|c|c|c|c|c|c|c|c}
\toprule
\multirow{3}{*}{Method} & \multicolumn{3}{c|}{KITTI} & \multicolumn{6}{c}{Waymo} \\
\cmidrule(r){2-10} 
& \multicolumn{3}{c|}{Car} & \multicolumn{2}{c|}{Veh.} & \multicolumn{2}{c|}{Ped.} & \multicolumn{2}{c}{Cyc.} \\
  & Easy & Mod. & Hard & L1 & L2 & L1 & L2 & L1 & L2  \\ 
\midrule
 Init \cite{deng2021voxel} \cite{shi2020pv} & 92.10 & 86.40 & 84.39 & 75.4 & 67.4 &72.0 & 63.7 & 65.9 & 63.4  \\
+ IoU Pred. & 93.40 & 87.23 & 85.50 & 75.4 & 67.7 & 72.4 & 64.2 & 66.0 & 63.8  \\
\: + MS Fusion& 96.01 & 88.97 & 88.31 & 75.8 & 68.0 & 72.7 & 64.7 & 66.5 & 64.0  \\
\bottomrule
\end{tabular}
}
\label{tab_component_comp_kitti}
\end{table}

\begin{table}[!t]
    \centering
    \caption{Comparison with previous methods employing multi-modal fusion or IoU joint prediction strategies, all implemented on the VoxelRCNN baseline. The results are reported in terms of 3D mAP for the Car class at 40\% recall.}
    \renewcommand\arraystretch{1}
    \resizebox{1.0\linewidth}{!}{
    \begin{tabular}{@{}@{\extracolsep{\fill}}!{\color{white}\vline}l|c|c|c @{}}
    \toprule
    \multirow{2}{*}{Method} & \multicolumn{3}{c}{$AP_{3D}$} \\ 
       & Easy & Mod. & Hard  \\ \midrule
       VoxelRCNN & 92.38 & 85.29 & 82.86  \\
       % \: + GoF, LoF (LoGoNet\cite{li2023logonet}) & 92.04 \textit{(-0.34)} & 85.04 \textit{(-0.25)} & 84.31 \textit{(+1.45)}  \\
       + VPF (VoPiNet\cite{wang2024vopifnet}) & - &  86.08 \textit{(+0.79)} & -    \\
        + NRConv (VirConv\cite{wu2023virtual}) & 93.36 \textit{(+0.98)} & 88.71 \textit{(+3.42)} & 85.83 \textit{(+2.97)} \\
       + 3D-GAF (SFD\cite{wu2022sparse}) &  95.47 \textit{(+3.09)} & 88.56 \textit{(+3.27)} & 85.74 \textit{(+2.88)}   \\
         \rowcolor{blue!10} + CMF-IOU (Ours) &  \textbf{96.01} \textit{(+3.63)} & \textbf{88.97} \textit{(+3.68)} & \textbf{88.31} \textit{(+5.45)} \\ \midrule
        VoxelRCNN &  92.38  & 85.29  & 82.86  \\
        + 3D IoU-Net \cite{li20203d} & 93.45 \textit{(+1.07)} & 85.71 \textit{(+0.42)} & 83.24 \textit{(+0.38)}  \\
        + From-Voxel2Point \cite{li2021voxel}& 93.00 \textit{(+0.62)} & 85.61 \textit{(+0.32)} & 83.43 \textit{(+0.57)} \\
        \rowcolor{blue!10} + CMF-IOU (Ours) & \textbf{93.68} \textit{(+1.3)} & \textbf{86.12} \textit{(+0.83)} & \textbf{83.97} \textit{(+1.11)} \\
         \bottomrule
    \end{tabular}
    }
    \label{tab_pre_component_comp}
\end{table}

\begin{table}[!t]
\scriptsize
\centering
\caption{Effectiveness of each part in the 3d convolution backbone. "Init" denotes the original VoxelRCNN backbone. "S2D" indicates the addition of the sparse-to-distant branch, and “ResVC” represents the integration of the residual view-consistency branch.}
\renewcommand\arraystretch{1}
\setlength{\tabcolsep}{1.3mm}{
\begin{tabular}{@{}@{\extracolsep{\fill}}!{\color{white}\vline}c|c|c|c|c|c|c|c @{}}
\toprule
\multirow{2}{*}{Init} & \multirow{2}{*}{S2D} & \multirow{2}{*}{ResVC} & \multicolumn{3}{c|}{AP$_{3D} (\%)$}   & \multirow{2}{*}{Params} & \multirow{2}{*}{RunTime}   \\ \cmidrule(r){4-6} 
&    &    & Easy &Mod.  & Hard &  &   \\ \midrule\checkmark &   &  & 94.78  & 88.45 &  85.99 & 14.8M & 26.5ms\\
\checkmark & \checkmark &  &  95.52& 88.61 & 86.09  & 15.4M & 28.1ms \\
\checkmark&   & \checkmark&  95.86& 88.93 & 88.14  & 15.2M & 33.7ms \\
\rowcolor{blue!10}\checkmark & \checkmark &\checkmark & \textbf{96.01} & \textbf{88.97} & \textbf{88.31} & 15.8M & 35.5ms\\ 
\bottomrule
\end{tabular}
}
\label{3b_backbone_ablation}
\end{table}

\begin{table}[!t]
\centering
\caption{ Performance under sensor noises with noise ratios ($\alpha_s$) of 1\%, 5\%, and 15\%. Results are reported in terms of 3D mAP in the moderate difficulty level.}
\renewcommand\arraystretch{1}
\resizebox{0.9\linewidth}{!}{
    \begin{tabular}{@{}@{\extracolsep{\fill}}!{\color{white}\vline}l|c|c|c|c|c|c @{}}
    \toprule
    Method & $\alpha_s$ & Car & Pedestrian & Cyclists & mAP & Dec. $\downarrow$ \\ \midrule
        VirConv & \multirow{2}{*}{/} & 88.18 & 62.52 & 74.42 & 75.04 & / \\
        CMF-IOU & & 88.48 & 65.13 & 75.90 & 76.50 & / \\ \midrule
        VirConv & \multirow{2}{*}{1\%} & 87.98 & 61.27 & 74.66 & 74.63 & 0.41 \\
        CMF-IOU & & 88.18 & 64.30 & 76.03 & 76.17 & 0.33 \\ \midrule
        VirConv & \multirow{2}{*}{5\%} & 85.92 & 55.51 & 68.60 & 70.01 & 5.03 \\
        CMF-IOU & & 85.98 & 62.90 & 71.20 & 73.36 & 3.14 \\ \midrule
        VirConv & \multirow{2}{*}{10\%} & 83.10 & 50.12 & 65.15 & 66.12 & 8.92\\
        CMF-IOU & & 83.39 & 59.25 & 69.18 & 70.61 & 5.89 \\
    \bottomrule
    \end{tabular}
    }
\label{tab_sensor_noise_comparison}
\end{table}

\begin{table}[!t]
\centering
\caption{ Performance under calibration errors with angular bias $\alpha_c$. Results are reported in terms of 3D mAP in the moderate difficulty level.}
\renewcommand\arraystretch{1}
\resizebox{0.9\linewidth}{!}{
\begin{tabular}{@{}@{\extracolsep{\fill}}!{\color{white}\vline}l|c|c|c|c|c|c @{}}
    \toprule
    Method & $\alpha_c$ & Car & Pedestrian & Cyclists & mAP & Dec. $\downarrow$ \\ \midrule
        VirConv & \multirow{2}{*}{/} & 88.18 & 62.52 & 74.42 & 75.04 & / \\
        CMF-IOU & & 88.48 & 65.13 & 75.90 & 76.50 & / \\ \midrule
        VirConv & \multirow{2}{*}{0.5°} & 82.75 & 63.03 & 72.29 & 72.69 & 2.35\\
        CMF-IOU & & 83.08 & 64.76 & 74.85 & 74.23 & 2.27 \\ \midrule
        VirConv & \multirow{2}{*}{1.0°} & 78.74 & 62.07 & 70.61 & 70.47 & 4.57 \\
        CMF-IOU & & 79.89 & 64.10 & 72.00 & 71.99 & 4.51 \\ \midrule
        VirConv & \multirow{2}{*}{2.0°} & 78.73 & 57.21 & 65.67 & 67.20 & 7.84\\
        CMF-IOU & & 78.82 & 58.18 & 68.80 & 68.60 & 7.90 \\
    \bottomrule
    \end{tabular}
    }
\label{tab_calibration_comparison}
\end{table}

\begin{table}[!t]
    \centering
    \caption{ Performance on our ResVC backbone and basic backbone in VoxelRCNN with or w/o the added depth noises.}
    \renewcommand\arraystretch{1}
    \resizebox{0.95\linewidth}{!}{
    \begin{tabular}{@{}@{\extracolsep{\fill}}!{\color{white}\vline}l|c|c|c|c|c|c @{}}
    \toprule
    Encoder & added noise & Car & Pedestrian & Cyclists & mAP & Dec. $\downarrow$ \\ \midrule
        VoxelRCNN \cite{deng2021voxel}& \multirow{2}{*}{\textit{w/o}}&  87.68 & 63.39 &  73.61  & 77.02 & / \\
        ResVC &  &  88.48 & 65.13 &  75.90 &  78.57 & / \\
         \midrule
        VoxelRCNN \cite{deng2021voxel}& \multirow{2}{*}{\textit{w/}} &  85.90& 62.20& 72.63 &  76.10 & 0.92 \\
        ResVC &  &  88.35 &  65.05 &  74.84 &  78.07 & 0.50 \\
    \bottomrule
    \end{tabular}
    }
    \label{tab_depth_noise_comparison}
\end{table}

\begin{figure}[!ht]
\centering
\begin{minipage}{0.24\textwidth}
\centering
\includegraphics[width=1\linewidth]{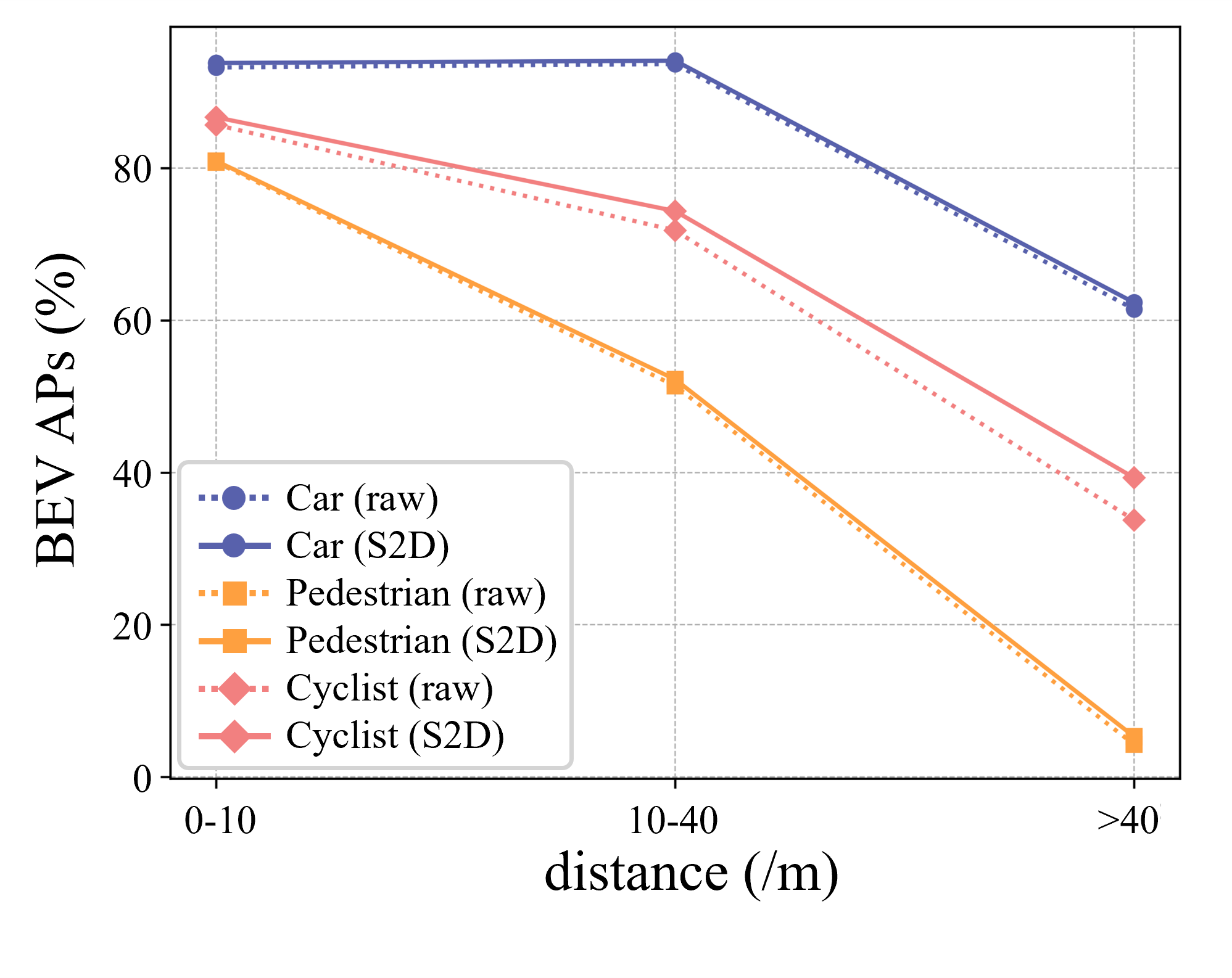}
\subfloat{\footnotesize (a)}
\end{minipage}
% \hspace{0.05cm}
\begin{minipage}{0.24\textwidth}
\centering
\includegraphics[width=1\linewidth]{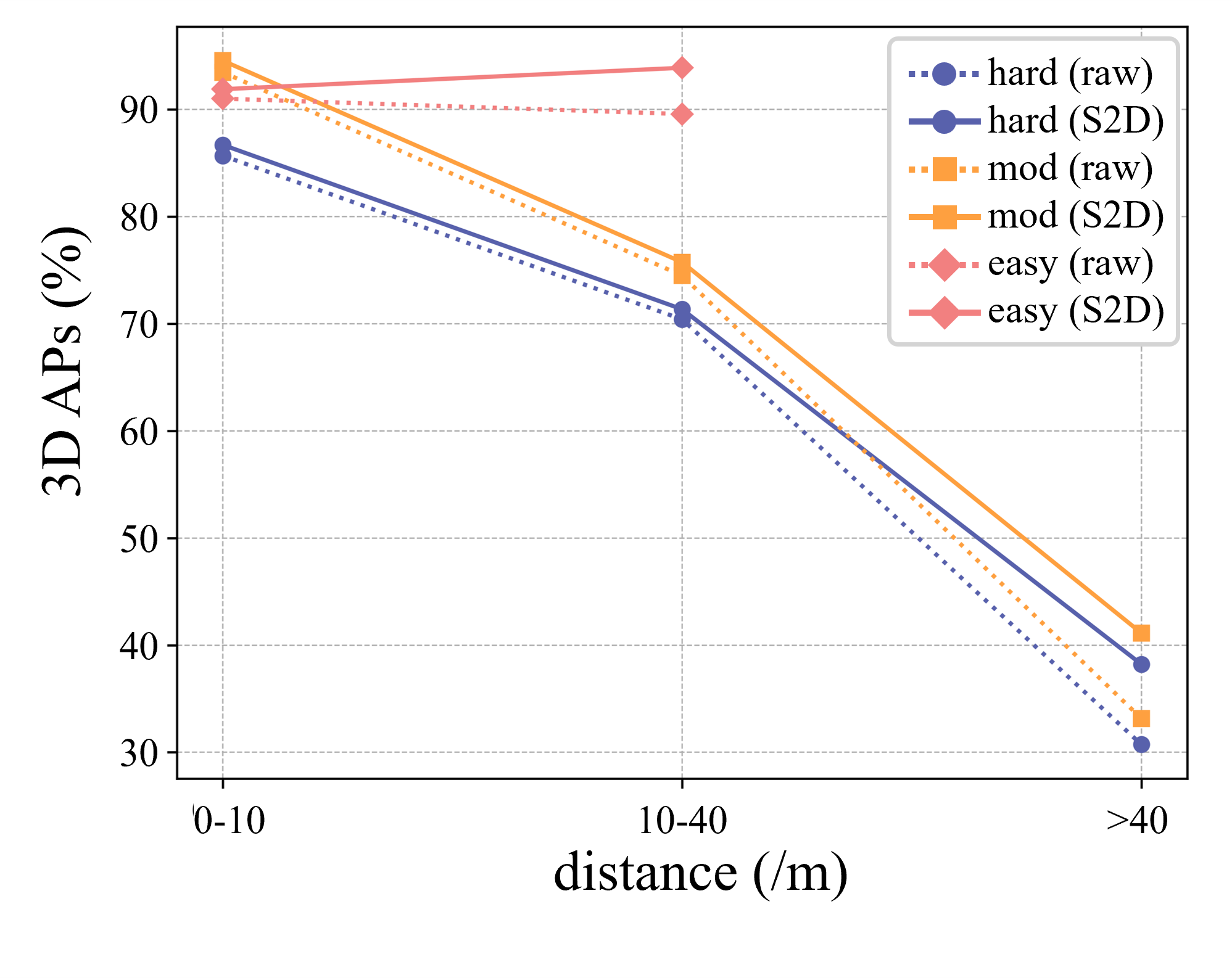}
\subfloat{\footnotesize (b)}
\end{minipage}

\caption{Improvement in APs across nearby, middle, and far distances. (a) BEV APs \textit{with} or \textit{without} the S2D branch across different categories. (b) 3D APs \textit{with} or \textit{without} the S2D branch across different difficulty levels in the cyclist category. }
\label{Fig:S2D branch}
\end{figure}

\begin{table}[!ht]
\scriptsize
\centering
\caption{Ablation of different ways in RoI features pooling. ``RV" represents raw voxel-aware pooling. ``PV" refers to the addition of pseudo voxel-aware pooling. ``RP" denotes the addition of raw point-aware pooling. ``PP" indicates the addition of pseudo point-aware pooling.}
\renewcommand\arraystretch{1}
\setlength{\tabcolsep}{1.3mm}{
\begin{tabular}{@{}@{\extracolsep{\fill}}!{\color{white}\vline}c|c|c|c|c|c|c|c|c @{}}
\toprule
\multirow{2}{*}{RV}& \multirow{2}{*}{PV}& \multirow{2}{*}{RP}& \multirow{2}{*}{PP} & \multicolumn{3}{c|}{AP$_{3D} (\%)$}   & \multirow{2}{*}{Params} & \multirow{2}{*}{RunTime}   \\ \cmidrule(r){5-7} 
&    &    &   & Easy &Mod.  & Hard &  &   \\ \midrule
\checkmark &   &  &  & 93.40 & 87.23&  85.50& 14.6M&  26.0ms\\
\checkmark & \checkmark &  & & 95.69& 88.12& 86.91& 15.2M&  29.4ms\\
\checkmark&   & \checkmark&  \checkmark &95.46& 88.22& 87.32& 15.3M& 30.2ms\\
\rowcolor{blue!10}\checkmark & \checkmark &\checkmark & \checkmark & \textbf{96.01} & \textbf{88.97} & \textbf{88.31} & 15.8M & 35.5ms\\ 
\bottomrule
\end{tabular}
}
\label{pooling_ways_ablation}
\end{table}

\begin{table}[!ht]
\scriptsize
\centering
\caption{Impact of the number of generated proposals after the RPN. Results are conducted in 3D and BEV APs calculated by 40 recall threshold for car class on the KITTI validation dataset. The initial IoU threshold of the generated proposals is from 0.5 to 1.0.}
\renewcommand\arraystretch{1}
\setlength{\tabcolsep}{1.3mm}{
\begin{tabular}{@{}@{\extracolsep{\fill}}!{\color{white}\vline}l|c|c|c|c|c|c|c @{}}
\toprule
\multirow{2}{*}{$N_{sp}$}  & \multicolumn{3}{c|}{AP$_{3D} (\%)$}   &  \multicolumn{3}{c|}{AP$_{BEV} (\%)$} & \multirow{2}{*}{RunTime}   \\ \cmidrule(r){2-7} 
& Easy & Mod.  & Hard & Easy & Mod.  & Hard &  \\ \midrule
10 & \textbf{96.59}& 88.79& 86.51& \textbf{96.47} & 93.83 & 91.55 &  34.6ms\\
30 & 96.46& 88.79& 86.99& 96.41& 93.87 & 91.58 &  34.3ms\\
80& 96.38& 88.81& 87.14& 96.33 & 93.93 & 91.67 & 34.8ms\\
\rowcolor{blue!10}100 &96.01& \textbf{88.97}& \textbf{88.31}& 96.30 & \textbf{93.97} & \textbf{91.69} & 35.5ms\\ 
160 & 96.42& 88.91& 86.66& 96.32 & 93.91 & 91.63 & 35.6ms\\ 
\bottomrule
\end{tabular}
}
\label{sampling_number}
\end{table}

\begin{table}[!ht]
\scriptsize
\centering
\caption{Impact of the IoU threshold on proposal generation. Results are reported in terms of 3D AP calculated at a 40 recall threshold for the car class on the KITTI validation dataset using our CMF-IoU framework. The number of generated proposals is set to 100.}
\renewcommand\arraystretch{1}
\setlength{\tabcolsep}{1.3mm}{
\begin{tabular}{@{}@{\extracolsep{\fill}}!{\color{white}\vline}c|c|c|c|c|c|c @{}}
\toprule
\multirow{2}{*}{${TS}_{1}$}& \multirow{2}{*}{${TS}_{2}$}& \multirow{2}{*}{${TS}_{3}$}& \multicolumn{3}{c|}{AP$_{3D} (\%)$}  & \multirow{2}{*}{RunTime}   \\ \cmidrule(r){4-6} 
&  &  & Easy & Mod.  & Hard &  \\ \midrule
\rowcolor{blue!10}$[0.5, 1.0)$& $[0.6, 1.0)$& $[0.7, 1.0)$& 96.01& \textbf{88.97}& \textbf{88.31}& 35.5ms\\
$[0.6, 1.0)$ & $[0.7, 1.0)$ & $[0.8, 1.0)$ & \textbf{96.36}& 88.73& 86.47& 35.4ms\\
$[0.7, 1.0)$ & $[0.8, 1.0)$ & $[0.9, 1.0)$ & 96.03& 88.58& 86.70&  35.4ms\\
$[0.8, 1.0)$ & $[0.9, 1.0)$ & $[0.95, 1.0)$ &  96.12& 88.84& 86.58& 35.3ms\\ 
\bottomrule
\end{tabular}
}
\label{sampling_threshold}
\end{table}

\begin{figure}[!ht] %%图1
	\centering  %插入的图片居中表示
	\includegraphics[width=1\linewidth]{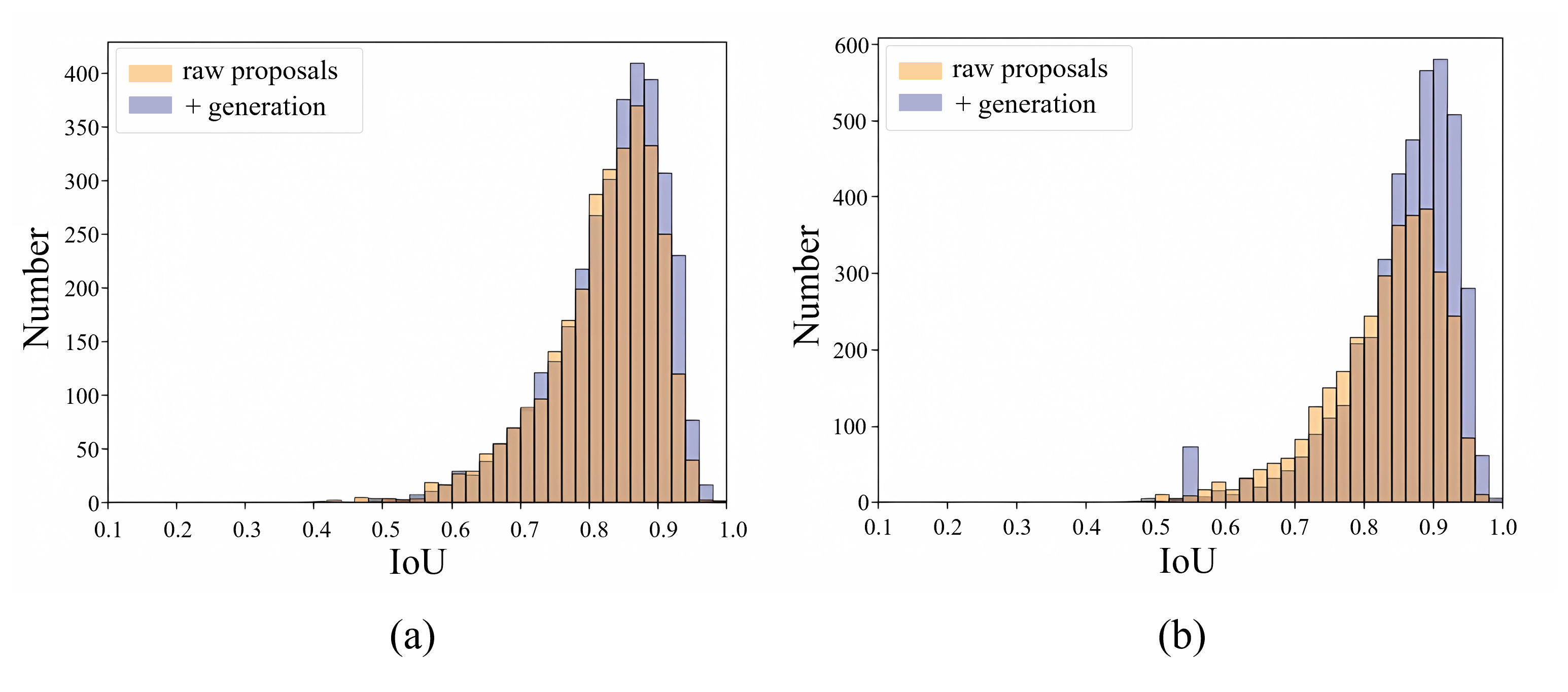}
	\caption{Number of predicted proposals in different IoU intervals. ``raw proposals'' represents the performance without IoU joint prediction, while the ``+ generation'' indicates results with IoU joint prediction. (a) Comparison before the refinement stage. (b) Comparison after the refinement stage.}
	\label{fig:IoU_number}  
\end{figure}

\begin{table}[!ht]
    \centering
    \caption{ Results of various IoU series metrics used for the IoU joint prediction branch on the KITTI and Waymo datasets.}
    \renewcommand\arraystretch{1}
    \resizebox{0.85\linewidth}{!}{
    \begin{tabular}{c|c|c|c|c|c|c}
    \toprule
     \multirow{2}{*}{Metrics} & \multicolumn{3}{c|}{KITTI} & \multicolumn{3}{c}{Waymo}  \\ 
     & Car & Ped. & Cyc. & Veh. & Ped. & Cyc. \\
    \midrule
            IoU &  88.48 & 65.13 & 75.90 & 68.01 & 64.74 & 64.02 \\
            GIoU & 88.47 & 65.06 & 74.88 & 67.36 & 64.22 & 63.19 \\
            DIoU & 88.22 & 64.51 & 75.24 & 67.19 & 64.14 & 63.47 \\
            CIoU & 89.54 & 65.03 & 76.48 & 68.36 & 64.64 & 63.70 \\
    \bottomrule
    \end{tabular}
    }
    \label{comp_metrics_kitti}
\end{table}

\begin{table}[!ht]
    \centering
    \caption{ Performance of different approaches on the NMS post-processing.}
    \renewcommand\arraystretch{1}
    \resizebox{0.85\linewidth}{!}{
    \begin{tabular}{c|c|c|c}
    \toprule
    Method & Car & Ped. & Cyc.  \\ \midrule
    Vanilla NMS &  88.48 & 65.13 & 75.90 \\
    Adaptive NMS Threshold &  88.41 & 65.26  & 75.98  \\
    Multi-Hypothesis Tracking & 87.62  & 64.20 & 75.04  \\
    \bottomrule
    \end{tabular}
    }
    \label{tab_nms_comp}
     \vspace{-0.5 cm}
\end{table}

\subsubsection{Robustness of the cross-view strategy in backbone}
To ensure the practicality and reliability of our model, we further evaluate the robustness under sensor or depth noises and calibration errors. In Table \ref{tab_sensor_noise_comparison}, we introduce coordinate noise to a portion of points (with a ratio of $\alpha_s$) during the reference stage. The performance degradation in our method is less than that in the compared model, VirConv \cite{wu2023virtual}, demonstrating our resilience to sensor disturbances. Additionally, to simulate calibration errors, we rotate all pseudo points by random angles within an upper bound defined by $\alpha_c$. Although both our method and VirConv experience performance drops, our approach consistently achieves better results, as reported in \ref{tab_calibration_comparison}. To further assess the robustness, we simulate realistic depth noise by introducing perturbations to 30\% of the pseudo points during inference. The results in Table \ref{tab_depth_noise_comparison} reveal that the VoxelRCNN \cite{deng2021voxel} backbone suffers from a substantial drop in performance due to its exclusive reliance on 3D geometric representation. In contrast, our method exhibits only a marginal degradation, showing its robustness in depth noise.CMF-IOU

\subsubsection{Prediction performance of the S2D branch at different distance}
To assess the instance information enhancement capability of our proposed S2D branch at long distances, we divide the evaluation samples into three parts based on the center distance of ground-truth boxes and evaluate our model on each part. The improvement in BEV APs across different categories with and without the S2D branch is illustrated in Fig.\ref{Fig:S2D branch}(a), where the overall performance amplification in far distance ($>40m$) object detection accuracy is higher than that in nearby distance ($<10 m$). Although the feature interaction in large instances (e.g. car) is less evident, the branch significantly enhances the features of small and remote instances, such as pedestrians and cyclists. Additionally, we compare the 3D APs of the cyclist category across different difficulty levels. As shown in Fig.\ref{Fig:S2D branch}(b), the improvements of moderate and hard difficulty at far distance achieve 7.97\% and 7.49\%, respectively, which is higher than the 1.08\% and 1.01\% improvements at nearby distance. It is worth noting that there are no easy difficulty level objects in distance $>40m$. The above results show that our S2D branch can refine the information in far and small objects to enhance the feature representation for better performance.

\subsubsection{Influence of different pooling strategies}
We also investigate the influence of different pooling strategies in the refinement head, including only raw voxel-aware pooling, the addition of pseudo voxel-aware pooling, and the addition of point-aware pooling. It can be observed in Table \ref{pooling_ways_ablation} that utilizing only the raw voxel features for grid pooling limits the performance due to the absence of more semantic and fine grained information, resulting in only 93.40\%, 87.23\% and 85.50\% in 3D APs. However, integrating voxel-aware pooling with pseudo voxels significantly enhances detection accuracy by supplementing the grid features with more details.

Meanwhile, the point-aware pooling assembling the regional and semantic information from irregular points can provide more precise 3D location and geometric optimization for each proposal. By combining various pooling methods, the final detection metrics can achieve 96.01\%, 88.97\% and 88.31\% in 3D APs, with improvements of 2.61\%, 1.74\%, and 2.81\%, respectively. These results demonstrate the necessity of integrating multiple features in the late stage for enhancing detection performance.

\begin{figure*}[!ht] %%图1
    \centering  % 插入的图片居中表示
    \begin{minipage}[b]{0.24\textwidth}
        \centering
        \includegraphics[width=1\linewidth]{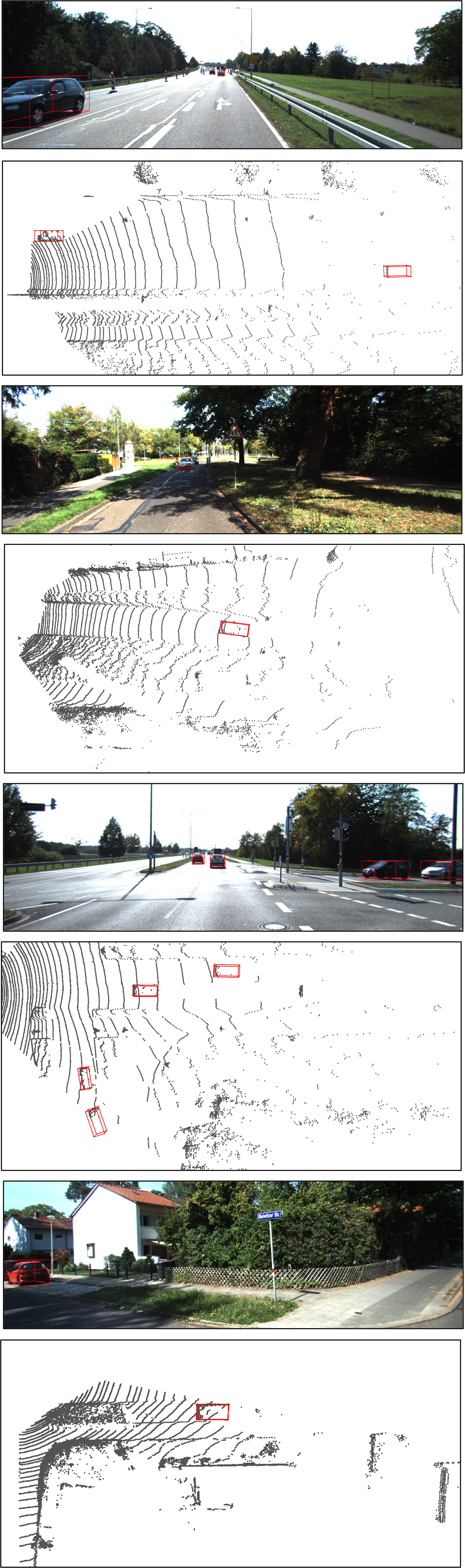}
        \subfloat{(a) GT}
    \end{minipage}
    \begin{minipage}[b]{0.24\textwidth}
        \centering
        \includegraphics[width=1\linewidth]{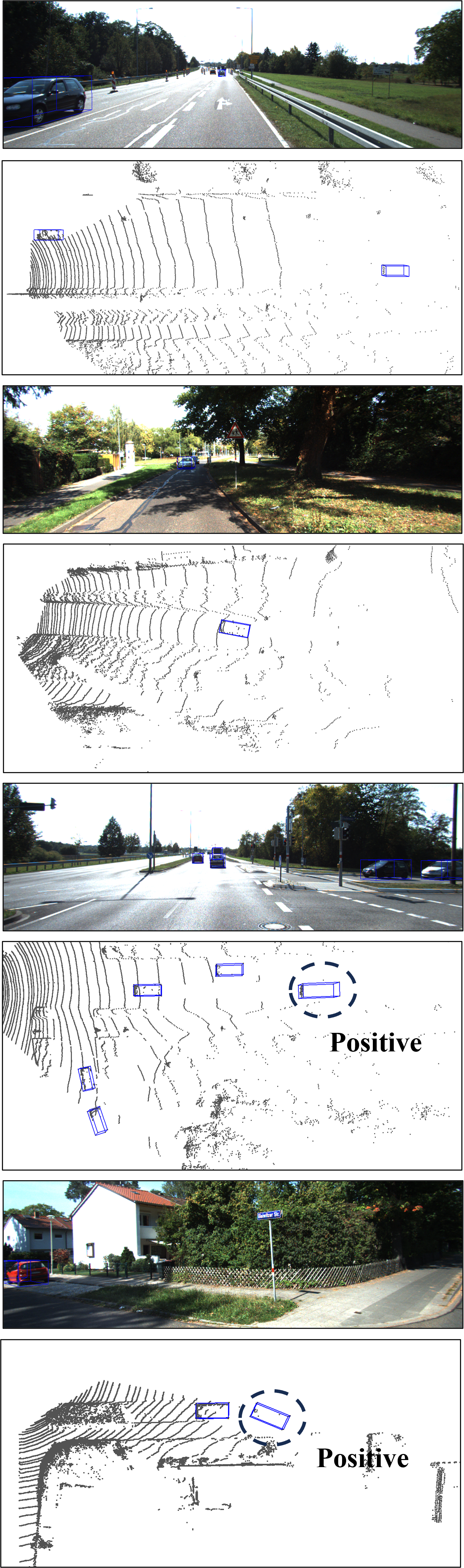}
        \subfloat{(b) CMF-IoU (Ours)}
    \end{minipage}
    \begin{minipage}[b]{0.24\textwidth}
        \centering
        \includegraphics[width=1\linewidth]{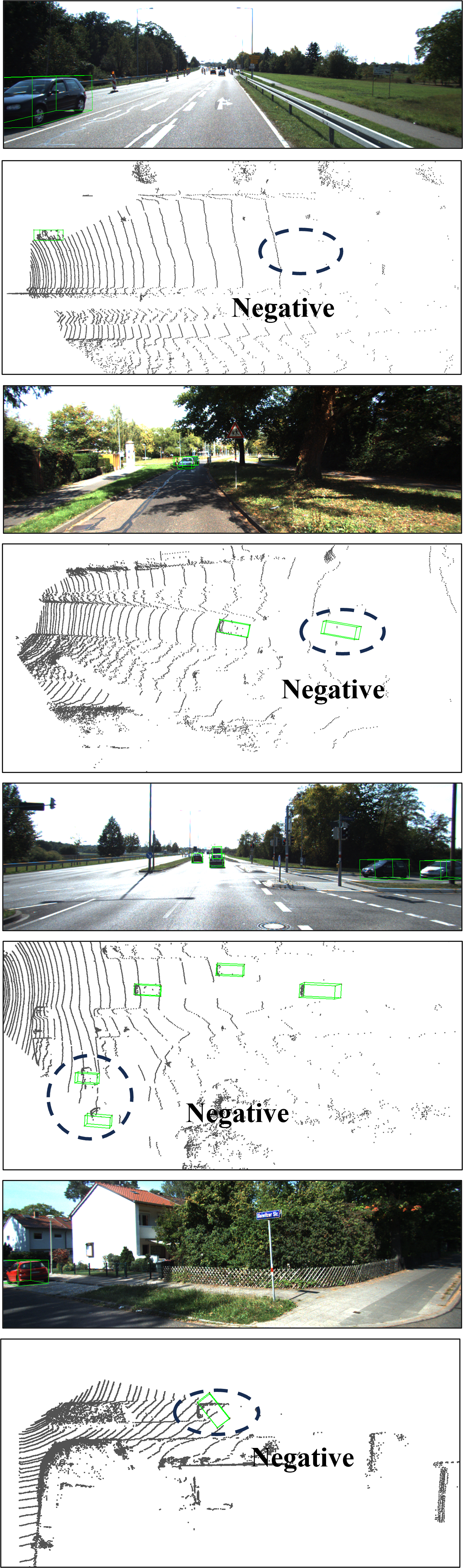}
        \subfloat{(c) VirConv}
    \end{minipage}
    \begin{minipage}[b]{0.24\textwidth}
        \centering
        \includegraphics[width=1\linewidth]{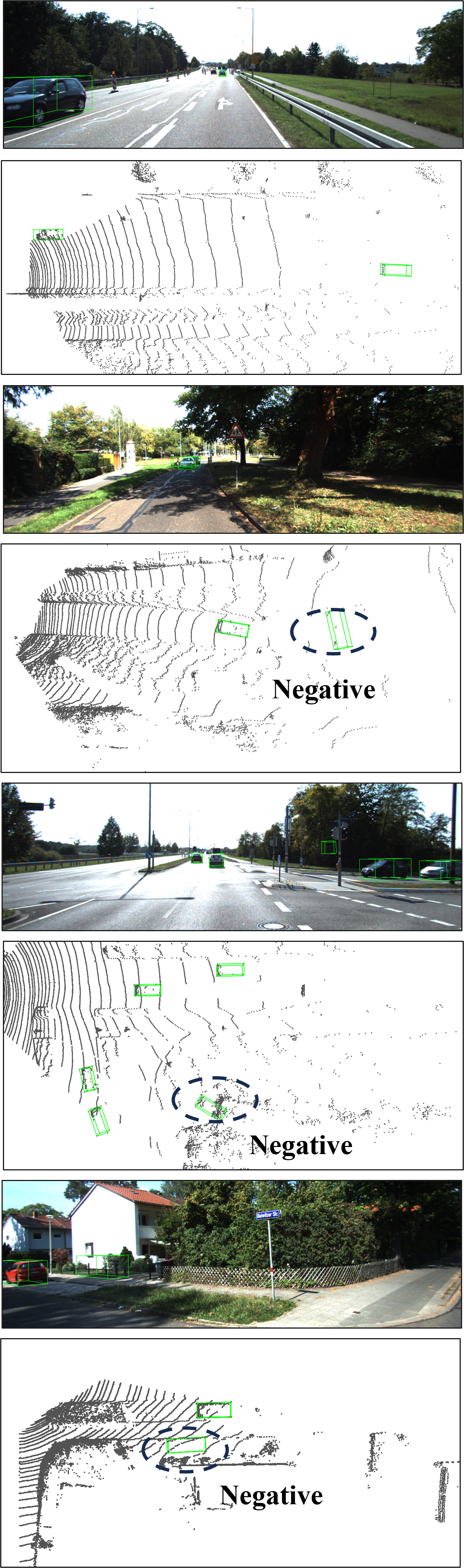}
        \subfloat{(d) VoxelRCNN}
    \end{minipage}
    \caption{Qualitative visualization results on the KITTI dataset. The upper rows show 3D detection results in image viewpoint, and the bottom rows show 3D detection results in 3D space.}
    \label{fig:visualization}
\end{figure*}

\subsubsection{Impact of the uniform RoIs generation}
We evaluate the performance under different hyperparameter settings during the uniform RoIs generation stage. Firstly, we randomly select the $N_{sp}$ proposals from the total $M\times N_g$ generated proposals in the entire scene. The number of proposals predicted from RPN is equal to $ 160 - N_{sp}$, ensuring that a total of $160$ RoIs are utilized for the IoU score prediction branch. As shown in Table \ref{sampling_number}, the 3D APs in three difficulty levels start at 96.59\%, 88.79\% and 86.51\%, respectively. As the sampled number increases  (from $10$ to $100$), the APs improve significantly in the moderate and hard levels, with gains of 0.18\% and 1.80\%. However, when all RoIs are generated ($N_{sp}=160$), the performance at the hard level declines, which is attributed to the domain gap between the generated RoIs during the training and the predicted RoIs during the validation. Thus, we select $100$ generated proposals for each scene as the default.

Secondly, we investigate the impact of different IoU thresholds on the proposal generation. As shown in Table \ref{sampling_threshold}, we define different IoU ranges in the initial iteration and increase the threshold in later iterations. It can be observed that when the average IoU threshold in each iteration is relatively lower (e.g. the first row in the table), the performance shows certain detection advantages for objects in moderate and hard difficulty levels. This suggests that the generated proposals with a lower IoU threshold can better match the RPN prediction for challenging instances, enabling the subsequent refinement head to focus more on optimizing these hard instances. We thus conduct the IoU range $[0.5, 1.0)$ as our default setting.

\subsubsection{Impact of the IoU joint prediction}
As shown in Fig. \ref{fig:IoU_number}, we visualize the number of proposals with different IoU values before and after the refinement in the validation. It can be observed that the IoU distribution of proposals before the refinement head is similar whether or not the IoU joint prediction branch is used. However, when the IoU joint prediction branch is employed, the final result after the refinement head shows a higher proportion of high-IoU proposals compared to the one without the IoU joint prediction branch. This indicates that our proposed IoU joint prediction method significantly increases the number of high-IoU proposals, which in turn enhances the accuracy of the reserved boxes during the post-processing stage. Consequently, the improved optimization in the regression task leads to an overall improvement in the final detection performance.

\subsubsection{Comparison of different IoU series loss functions}
To investigate the impact of the different loss functions used in the IoU joint prediction branch, we replace the original loss with GIOU \cite{rezatofighi2019generalized}, DIOU, and CIOU \cite{zheng2020distance}. The calculation of these loss functions is similar to the IoU loss function used in 3D detection. As shown in \ref{comp_metrics_kitti}, the models trained with the CIOU and standard IoU loss outperform those applying GIoU or DIoU. However, the distinct between the CIoU and vanilla IoU is relatively minor and can be considered negligible. Therefore, we utilize the IoU as our default metric.

\subsubsection{Comparison of the post-processing strategies}
The NMS could inadvertently suppress proposals belonging to distinct instances. This issue leads to deficient detection boxes and reduced recall, which remains an inherent limitation of traditional NMS strategies. Therefore, we employ and compare the effectiveness of adaptive NMS threshold (ANT) \cite{liu2019adaptive, lee2023multi} and multi-hypothesis tracking (MHT) \cite{kim2015multiple, sheng2018iterative} approaches in the post-processing stage. The result in \ref{tab_nms_comp} illustrates that the ANT yields modest improvements for smaller and denser instances such as pedestrians and cyclists. However, the MHT shows negligible benefit across all categories, which is attributed to the uncertainty in the feature distributions of the proposals. 

\subsection{Visualization}
Fig. \ref{fig:visualization} presents a qualitative comparison of our CMF-IoU against the state-of-the-art method VirConv \cite{wu2023virtual} and our baseline model VoxelRCNN \cite{deng2021voxel} on the specific scenes from the KITTI dataset. As illustrated in the first group of examples, our method accurately detects distant vehicle, whereas the VirConv fails to capture them. In the next group, both VirConv and VoxelRCNN misclassify the road obstacles as vehicles, while our method produces correct detection results. In the third and fourth groups, the VirConv predicts incorrect yaw angles for the boxes and VoxelRCNN regards the shadowed regions in the image as vehicles. Meanwhile, our method can recognize the heavily occluded or truncated objects that are not annotated in the ground truth. These visualizations demonstrate that CMF-IoU achieves higher recall on distant and challenging targets and exhibits better generalization in complex scenes, ensuring reliable detection performance.

\section{Conclusion}
In this paper, we introduce a novel multi-modal 3D object detector, CMF-IoU. Unlike previous single-stage fusion approaches, our method achieves promising performance by effectively fusing multi-modal information across the early, middle, and late stages. Specifically, we address dimensional misalignment issue by leveraging depth estimation to directly project image information into the 3D space. In the middle stage, we mitigate depth errors in pseudo points and the sparsity of LiDAR points by employing a bilateral cross-view enhanced backbone, rather than directly merging them for encoding. The features encoded by both branches are then aggregated in the BEV space. For late stage fusion in the refinement phase, we design an iterative voxel-point aware fine grained pooling module to extract spatial and textual information in the proposals. The features within the proposals across different iterations are further interacted through a cross-attention mechanism. Finally, we propose an IoU joint prediction branch to optimize the metric by balancing both IoU and classification scores during the NMS process. Extensive comparison and ablation experiments on various 3D detection datasets demonstrate the superior performance of our methods.

\ifCLASSOPTIONcaptionsoff
  \newpage
\fi

\bibliographystyle{IEEEtran}
\bibliography{mybibfile}

\begin{thebibliography}{10}
\providecommand{\url}[1]{#1}
\csname url@samestyle\endcsname
\providecommand{\newblock}{\relax}
\providecommand{\bibinfo}[2]{#2}
\providecommand{\BIBentrySTDinterwordspacing}{\spaceskip=0pt\relax}
\providecommand{\BIBentryALTinterwordstretchfactor}{4}
\providecommand{\BIBentryALTinterwordspacing}{\spaceskip=\fontdimen2\font plus
\BIBentryALTinterwordstretchfactor\fontdimen3\font minus \fontdimen4\font\relax}
\providecommand{\BIBforeignlanguage}[2]{{%
\expandafter\ifx\csname l@#1\endcsname\relax
\typeout{** WARNING: IEEEtran.bst: No hyphenation pattern has been}%
\typeout{** loaded for the language `#1'. Using the pattern for}%
\typeout{** the default language instead.}%
\else
\language=\csname l@#1\endcsname
\fi
#2}}
\providecommand{\BIBdecl}{\relax}
\BIBdecl

\bibitem{mao20233d}
J.~Mao, S.~Shi, X.~Wang, and H.~Li, ``3d object detection for autonomous driving: A comprehensive survey,'' \emph{International Journal of Computer Vision}, vol. 131, no.~8, pp. 1909--1963, 2023.

\bibitem{qi2017pointnet}
C.~R. Qi, H.~Su, K.~Mo, and L.~J. Guibas, ``Pointnet: Deep learning on point sets for 3d classification and segmentation,'' in \emph{Proceedings of the IEEE conference on computer vision and pattern recognition}, 2017, pp. 652--660.

\bibitem{qi2017pointnet++}
C.~R. Qi, L.~Yi, H.~Su, and L.~J. Guibas, ``Pointnet++: Deep hierarchical feature learning on point sets in a metric space,'' \emph{Advances in neural information processing systems}, vol.~30, 2017.

\bibitem{liang2024pointmamba}
D.~Liang, X.~Zhou, X.~Wang, X.~Zhu, W.~Xu, Z.~Zou, X.~Ye, and X.~Bai, ``Pointmamba: A simple state space model for point cloud analysis,'' \emph{arXiv preprint arXiv:2402.10739}, 2024.

\bibitem{zhou2018voxelnet}
Y.~Zhou and O.~Tuzel, ``Voxelnet: End-to-end learning for point cloud based 3d object detection,'' in \emph{Proceedings of the IEEE conference on computer vision and pattern recognition}, 2018, pp. 4490--4499.

\bibitem{shi2020pv}
S.~Shi, C.~Guo, L.~Jiang, Z.~Wang, J.~Shi, X.~Wang, and H.~Li, ``Pv-rcnn: Point-voxel feature set abstraction for 3d object detection,'' in \emph{Proceedings of the IEEE/CVF conference on computer vision and pattern recognition}, 2020, pp. 10\,529--10\,538.

\bibitem{chen2017multi}
X.~Chen, H.~Ma, J.~Wan, B.~Li, and T.~Xia, ``Multi-view 3d object detection network for autonomous driving,'' in \emph{Proceedings of the IEEE conference on Computer Vision and Pattern Recognition}, 2017, pp. 1907--1915.

\bibitem{li2022bevformer}
Z.~Li, W.~Wang, H.~Li, E.~Xie, C.~Sima, T.~Lu, Y.~Qiao, and J.~Dai, ``Bevformer: Learning bird’s-eye-view representation from multi-camera images via spatiotemporal transformers,'' in \emph{European conference on computer vision}.\hskip 1em plus 0.5em minus 0.4em\relax Springer, 2022, pp. 1--18.

\bibitem{wang2022detr3d}
Y.~Wang, V.~C. Guizilini, T.~Zhang, Y.~Wang, H.~Zhao, and J.~Solomon, ``Detr3d: 3d object detection from multi-view images via 3d-to-2d queries,'' in \emph{Conference on Robot Learning}.\hskip 1em plus 0.5em minus 0.4em\relax PMLR, 2022, pp. 180--191.

\bibitem{liu2023bevfusion}
Z.~Liu, H.~Tang, A.~Amini, X.~Yang, H.~Mao, D.~L. Rus, and S.~Han, ``Bevfusion: Multi-task multi-sensor fusion with unified bird's-eye view representation,'' in \emph{2023 IEEE international conference on robotics and automation (ICRA)}.\hskip 1em plus 0.5em minus 0.4em\relax IEEE, 2023, pp. 2774--2781.

\bibitem{liang2022bevfusion}
T.~Liang, H.~Xie, K.~Yu, Z.~Xia, Z.~Lin, Y.~Wang, T.~Tang, B.~Wang, and Z.~Tang, ``Bevfusion: A simple and robust lidar-camera fusion framework,'' \emph{Advances in Neural Information Processing Systems}, vol.~35, pp. 10\,421--10\,434, 2022.

\bibitem{wu2023virtual}
H.~Wu, C.~Wen, S.~Shi, X.~Li, and C.~Wang, ``Virtual sparse convolution for multimodal 3d object detection,'' in \emph{Proceedings of the IEEE/CVF Conference on Computer Vision and Pattern Recognition}, 2023, pp. 21\,653--21\,662.

\bibitem{bai2022transfusion}
X.~Bai, Z.~Hu, X.~Zhu, Q.~Huang, Y.~Chen, H.~Fu, and C.-L. Tai, ``Transfusion: Robust lidar-camera fusion for 3d object detection with transformers,'' in \emph{Proceedings of the IEEE/CVF conference on computer vision and pattern recognition}, 2022, pp. 1090--1099.

\bibitem{yan2023cross}
J.~Yan, Y.~Liu, J.~Sun, F.~Jia, S.~Li, T.~Wang, and X.~Zhang, ``Cross modal transformer: Towards fast and robust 3d object detection,'' in \emph{Proceedings of the IEEE/CVF International Conference on Computer Vision}, 2023, pp. 18\,268--18\,278.

\bibitem{vora2020pointpainting}
S.~Vora, A.~H. Lang, B.~Helou, and O.~Beijbom, ``Pointpainting: Sequential fusion for 3d object detection,'' in \emph{Proceedings of the IEEE/CVF conference on computer vision and pattern recognition}, 2020, pp. 4604--4612.

\bibitem{huang2020epnet}
T.~Huang, Z.~Liu, X.~Chen, and X.~Bai, ``Epnet: Enhancing point features with image semantics for 3d object detection,'' in \emph{Computer Vision--ECCV 2020: 16th European Conference, Glasgow, UK, August 23--28, 2020, Proceedings, Part XV 16}.\hskip 1em plus 0.5em minus 0.4em\relax Springer, 2020, pp. 35--52.

\bibitem{yin2021multimodal}
T.~Yin, X.~Zhou, and P.~Kr{\"a}henb{\"u}hl, ``Multimodal virtual point 3d detection,'' \emph{Advances in Neural Information Processing Systems}, vol.~34, pp. 16\,494--16\,507, 2021.

\bibitem{wu2022sparse}
X.~Wu, L.~Peng, H.~Yang, L.~Xie, C.~Huang, C.~Deng, H.~Liu, and D.~Cai, ``Sparse fuse dense: Towards high quality 3d detection with depth completion,'' in \emph{Proceedings of the IEEE/CVF conference on computer vision and pattern recognition}, 2022, pp. 5418--5427.

\bibitem{deng2021voxel}
J.~Deng, S.~Shi, P.~Li, W.~Zhou, Y.~Zhang, and H.~Li, ``Voxel r-cnn: Towards high performance voxel-based 3d object detection,'' in \emph{Proceedings of the AAAI conference on artificial intelligence}, vol.~35, no.~2, 2021, pp. 1201--1209.

\bibitem{li2022spatial}
Z.~Li, Y.~Yao, Z.~Quan, J.~Xie, and W.~Yang, ``Spatial information enhancement network for 3d object detection from point cloud,'' \emph{Pattern Recognition}, vol. 128, p. 108684, 2022.

\bibitem{li20203d}
J.~Li, S.~Luo, Z.~Zhu, H.~Dai, A.~S. Krylov, Y.~Ding, and L.~Shao, ``3d iou-net: Iou guided 3d object detector for point clouds,'' \emph{arXiv preprint arXiv:2004.04962}, 2020.

\bibitem{zhu2021iou}
L.~Zhu, Z.~Xie, L.~Liu, B.~Tao, and W.~Tao, ``Iou-uniform r-cnn: Breaking through the limitations of rpn,'' \emph{Pattern Recognition}, vol. 112, p. 107816, 2021.

\bibitem{qiao2024local}
R.~Qiao, H.~Ji, Z.~Zhu, and W.~Zhang, ``Local-to-global semantic learning for multi-view 3d object detection from point cloud,'' \emph{IEEE Transactions on Circuits and Systems for Video Technology}, 2024.

\bibitem{shi2019pointrcnn}
S.~Shi, X.~Wang, and H.~Li, ``Pointrcnn: 3d object proposal generation and detection from point cloud,'' in \emph{Proceedings of the IEEE/CVF conference on computer vision and pattern recognition}, 2019, pp. 770--779.

\bibitem{tian2023medoidsformer}
X.~Tian, M.~Yang, Q.~Yu, J.~Yong, and D.~Xu, ``Medoidsformer: A strong 3d object detection backbone by exploiting interaction with adjacent medoid tokens,'' \emph{IEEE Transactions on Circuits and Systems for Video Technology}, vol.~33, no.~10, pp. 5844--5854, 2023.

\bibitem{shi2020points}
S.~Shi, Z.~Wang, J.~Shi, X.~Wang, and H.~Li, ``From points to parts: 3d object detection from point cloud with part-aware and part-aggregation network,'' \emph{IEEE transactions on pattern analysis and machine intelligence}, vol.~43, no.~8, pp. 2647--2664, 2020.

\bibitem{sheng2021improving}
H.~Sheng, S.~Cai, Y.~Liu, B.~Deng, J.~Huang, X.-S. Hua, and M.-J. Zhao, ``Improving 3d object detection with channel-wise transformer,'' in \emph{Proceedings of the IEEE/CVF International Conference on Computer Vision}, 2021, pp. 2743--2752.

\bibitem{wang2023long}
C.~Wang, J.~Deng, J.~He, T.~Zhang, Z.~Zhang, and Y.~Zhang, ``Long-short range adaptive transformer with dynamic sampling for 3d object detection,'' \emph{IEEE Transactions on Circuits and Systems for Video Technology}, vol.~33, no.~12, pp. 7616--7629, 2023.

\bibitem{luo2021m3dssd}
S.~Luo, H.~Dai, L.~Shao, and Y.~Ding, ``M3dssd: Monocular 3d single stage object detector,'' in \emph{Proceedings of the IEEE/CVF Conference on Computer Vision and Pattern Recognition}, 2021, pp. 6145--6154.

\bibitem{huang2023tri}
Y.~Huang, W.~Zheng, Y.~Zhang, J.~Zhou, and J.~Lu, ``Tri-perspective view for vision-based 3d semantic occupancy prediction,'' in \emph{Proceedings of the IEEE/CVF conference on computer vision and pattern recognition}, 2023, pp. 9223--9232.

\bibitem{gao2022esgn}
A.~Gao, Y.~Pang, J.~Nie, Z.~Shao, J.~Cao, Y.~Guo, and X.~Li, ``Esgn: Efficient stereo geometry network for fast 3d object detection,'' \emph{IEEE Transactions on Circuits and Systems for Video Technology}, 2022.

\bibitem{huang2021bevdet}
J.~Huang, G.~Huang, Z.~Zhu, Y.~Ye, and D.~Du, ``Bevdet: High-performance multi-camera 3d object detection in bird-eye-view,'' \emph{arXiv preprint arXiv:2112.11790}, 2021.

\bibitem{philion2020lift}
J.~Philion and S.~Fidler, ``Lift, splat, shoot: Encoding images from arbitrary camera rigs by implicitly unprojecting to 3d,'' in \emph{Computer Vision--ECCV 2020: 16th European Conference, Glasgow, UK, August 23--28, 2020, Proceedings, Part XIV 16}.\hskip 1em plus 0.5em minus 0.4em\relax Springer, 2020, pp. 194--210.

\bibitem{wang2022sts}
Z.~Wang, C.~Min, Z.~Ge, Y.~Li, Z.~Li, H.~Yang, and D.~Huang, ``Sts: Surround-view temporal stereo for multi-view 3d detection,'' \emph{arXiv preprint arXiv:2208.10145}, 2022.

\bibitem{song2023graphalign}
Z.~Song, H.~Wei, L.~Bai, L.~Yang, and C.~Jia, ``Graphalign: Enhancing accurate feature alignment by graph matching for multi-modal 3d object detection,'' in \emph{Proceedings of the IEEE/CVF International Conference on Computer Vision}, 2023, pp. 3358--3369.

\bibitem{jiang2018acquisition}
B.~Jiang, R.~Luo, J.~Mao, T.~Xiao, and Y.~Jiang, ``Acquisition of localization confidence for accurate object detection,'' in \emph{Proceedings of the European conference on computer vision (ECCV)}, 2018, pp. 784--799.

\bibitem{li2021voxel}
J.~Li, H.~Dai, L.~Shao, and Y.~Ding, ``From voxel to point: Iou-guided 3d object detection for point cloud with voxel-to-point decoder,'' in \emph{Proceedings of the 29th ACM International Conference on Multimedia}, 2021, pp. 4622--4631.

\bibitem{wu2022casa}
H.~Wu, J.~Deng, C.~Wen, X.~Li, C.~Wang, and J.~Li, ``Casa: A cascade attention network for 3-d object detection from lidar point clouds,'' \emph{IEEE Transactions on Geoscience and Remote Sensing}, vol.~60, pp. 1--11, 2022.

\bibitem{balta2018fast}
H.~Balta, J.~Velagic, W.~Bosschaerts, G.~De~Cubber, and B.~Siciliano, ``Fast statistical outlier removal based method for large 3d point clouds of outdoor environments,'' \emph{IFAC-PapersOnLine}, vol.~51, no.~22, pp. 348--353, 2018.

\bibitem{geiger2012we}
A.~Geiger, P.~Lenz, and R.~Urtasun, ``Are we ready for autonomous driving? the kitti vision benchmark suite,'' in \emph{2012 IEEE conference on computer vision and pattern recognition}.\hskip 1em plus 0.5em minus 0.4em\relax IEEE, 2012, pp. 3354--3361.

\bibitem{caesar2020nuscenes}
H.~Caesar, V.~Bankiti, A.~H. Lang, S.~Vora, V.~E. Liong, Q.~Xu, A.~Krishnan, Y.~Pan, G.~Baldan, and O.~Beijbom, ``nuscenes: A multimodal dataset for autonomous driving,'' in \emph{Proceedings of the IEEE/CVF conference on computer vision and pattern recognition}, 2020, pp. 11\,621--11\,631.

\bibitem{sun2020scalability}
P.~Sun, H.~Kretzschmar, X.~Dotiwalla, A.~Chouard, V.~Patnaik, P.~Tsui, J.~Guo, Y.~Zhou, Y.~Chai, B.~Caine \emph{et~al.}, ``Scalability in perception for autonomous driving: Waymo open dataset,'' in \emph{Proceedings of the IEEE/CVF conference on computer vision and pattern recognition}, 2020, pp. 2446--2454.

\bibitem{hu2021penet}
M.~Hu, S.~Wang, B.~Li, S.~Ning, L.~Fan, and X.~Gong, ``Penet: Towards precise and efficient image guided depth completion,'' in \emph{2021 IEEE International Conference on Robotics and Automation (ICRA)}.\hskip 1em plus 0.5em minus 0.4em\relax IEEE, 2021, pp. 13\,656--13\,662.

\bibitem{chen2023voxelnext}
Y.~Chen, J.~Liu, X.~Zhang, X.~Qi, and J.~Jia, ``Voxelnext: Fully sparse voxelnet for 3d object detection and tracking,'' in \emph{Proceedings of the IEEE/CVF Conference on Computer Vision and Pattern Recognition}, 2023, pp. 21\,674--21\,683.

\bibitem{zhang2024cvpr}
G.~Zhang, J.~Chen, G.~Gao, J.~Li, S.~Liu, and X.~Hu, ``{SAFDNet}: A simple and effective network for fully sparse 3d object detection,'' in \emph{Proceedings of the IEEE/CVF Conference on Computer Vision and Pattern Recognition (CVPR)}, June 2024, pp. 14\,477--14\,486.

\bibitem{yin2021center}
T.~Yin, X.~Zhou, and P.~Krahenbuhl, ``Center-based 3d object detection and tracking,'' in \emph{Proceedings of the IEEE/CVF conference on computer vision and pattern recognition}, 2021, pp. 11\,784--11\,793.

\bibitem{chen2020dsgn}
Y.~Chen, S.~Liu, X.~Shen, and J.~Jia, ``Dsgn: Deep stereo geometry network for 3d object detection,'' \emph{Proceedings of the IEEE Conference on Computer Vision and Pattern Recognition}, 2020.

\bibitem{yolostereo3d}
Y.~Liu, L.~Wang, and M.~Liu, ``Yolostereo3d: A step back to 2d for efficient stereo 3d detection,'' in \emph{2021 IEEE International Conference on Robotics and Automation (ICRA)}, 2021, pp. 13\,018--13\,024.

\bibitem{lang2019pointpillars}
A.~H. Lang, S.~Vora, H.~Caesar, L.~Zhou, J.~Yang, and O.~Beijbom, ``Pointpillars: Fast encoders for object detection from point clouds,'' in \emph{Proceedings of the IEEE/CVF conference on computer vision and pattern recognition}, 2019, pp. 12\,697--12\,705.

\bibitem{wang2024vopifnet}
C.-H. Wang, H.-W. Chen, Y.~Chen, P.-Y. Hsiao, and L.-C. Fu, ``Vopifnet: Voxel-pixel fusion network for multi-class 3d object detection,'' \emph{IEEE Transactions on Intelligent Transportation Systems}, 2024.

\bibitem{song2023graphalign++}
Z.~Song, C.~Jia, L.~Yang, H.~Wei, and L.~Liu, ``Graphalign++: An accurate feature alignment by graph matching for multi-modal 3d object detection,'' \emph{IEEE Transactions on Circuits and Systems for Video Technology}, vol.~34, no.~4, pp. 2619--2632, 2023.

\bibitem{li2023logonet}
X.~Li, T.~Ma, Y.~Hou, B.~Shi, Y.~Yang, Y.~Liu, X.~Wu, Q.~Chen, Y.~Li, Y.~Qiao \emph{et~al.}, ``Logonet: Towards accurate 3d object detection with local-to-global cross-modal fusion,'' in \emph{Proceedings of the IEEE/CVF Conference on Computer Vision and Pattern Recognition}, 2023, pp. 17\,524--17\,534.

\bibitem{wu2023transformation}
H.~Wu, C.~Wen, W.~Li, X.~Li, R.~Yang, and C.~Wang, ``Transformation-equivariant 3d object detection for autonomous driving,'' in \emph{Proceedings of the AAAI Conference on Artificial Intelligence}, vol.~37, no.~3, 2023, pp. 2795--2802.

\bibitem{zheng2022boosting}
W.~Zheng, M.~Hong, L.~Jiang, and C.-W. Fu, ``Boosting 3d object detection by simulating multimodality on point clouds,'' in \emph{Proceedings of the IEEE/CVF Conference on Computer Vision and Pattern Recognition}, 2022, pp. 13\,638--13\,647.

\bibitem{hu2022afdetv2}
Y.~Hu, Z.~Ding, R.~Ge, W.~Shao, L.~Huang, K.~Li, and Q.~Liu, ``Afdetv2: Rethinking the necessity of the second stage for object detection from point clouds,'' in \emph{Proceedings of the AAAI Conference on Artificial Intelligence}, vol.~36, no.~1, 2022, pp. 969--979.

\bibitem{deng2022vista}
S.~Deng, Z.~Liang, L.~Sun, and K.~Jia, ``Vista: Boosting 3d object detection via dual cross-view spatial attention,'' in \emph{Proceedings of the IEEE/CVF conference on computer vision and pattern recognition}, 2022, pp. 8448--8457.

\bibitem{chen2022autoalign}
Z.~Chen, Z.~Li, S.~Zhang, L.~Fang, Q.~Jiang, F.~Zhao, B.~Zhou, and H.~Zhao, ``Autoalign: Pixel-instance feature aggregation for multi-modal 3d object detection,'' \emph{arXiv preprint arXiv:2201.06493}, 2022.

\bibitem{chen2022focal}
Y.~Chen, Y.~Li, X.~Zhang, J.~Sun, and J.~Jia, ``Focal sparse convolutional networks for 3d object detection,'' in \emph{Proceedings of the IEEE/CVF Conference on Computer Vision and Pattern Recognition}, 2022, pp. 5428--5437.

\bibitem{focalformer3d}
Y.~Chen, Z.~Yu, Y.~Chen, S.~Lan, A.~Anandkumar, J.~Jia, and J.~M. Alvarez, ``Focalformer3d: Focusing on hard instance for 3d object detection,'' 2023.

\bibitem{wang2023dsvt}
H.~Wang, C.~Shi, S.~Shi, M.~Lei, S.~Wang, D.~He, B.~Schiele, and L.~Wang, ``Dsvt: Dynamic sparse voxel transformer with rotated sets,'' in \emph{CVPR}, 2023.

\bibitem{zhang2023hednet}
G.~Zhang, J.~Chen, G.~Gao, J.~Li, and X.~Hu, ``{HEDNet}: A hierarchical encoder-decoder network for 3d object detection in point clouds,'' in \emph{Thirty-seventh Conference on Neural Information Processing Systems (NeurIPS)}, 2023.

\bibitem{chen2022futr3d}
X.~Chen, T.~Zhang, Y.~Wang, Y.~Wang, and H.~Zhao, ``Futr3d: A unified sensor fusion framework for 3d detection,'' \emph{arXiv preprint arXiv:2203.10642}, 2022.

\bibitem{rezatofighi2019generalized}
H.~Rezatofighi, N.~Tsoi, J.~Gwak, A.~Sadeghian, I.~Reid, and S.~Savarese, ``Generalized intersection over union: A metric and a loss for bounding box regression,'' in \emph{Proceedings of the IEEE/CVF conference on computer vision and pattern recognition}, 2019, pp. 658--666.

\bibitem{zheng2020distance}
Z.~Zheng, P.~Wang, W.~Liu, J.~Li, R.~Ye, and D.~Ren, ``Distance-iou loss: Faster and better learning for bounding box regression,'' in \emph{Proceedings of the AAAI conference on artificial intelligence}, vol.~34, no.~07, 2020, pp. 12\,993--13\,000.

\bibitem{liu2019adaptive}
S.~Liu, D.~Huang, and Y.~Wang, ``Adaptive nms: Refining pedestrian detection in a crowd,'' in \emph{Proceedings of the IEEE/CVF conference on computer vision and pattern recognition}, 2019, pp. 6459--6468.

\bibitem{lee2023multi}
Y.~LEE, S.~Makonin, and K.~Noh, ``Multi-class object detection using adaptive non-maximum suppression in dense images,'' \emph{Authorea Preprints}, 2023.

\bibitem{kim2015multiple}
C.~Kim, F.~Li, A.~Ciptadi, and J.~M. Rehg, ``Multiple hypothesis tracking revisited,'' in \emph{Proceedings of the IEEE international conference on computer vision}, 2015, pp. 4696--4704.

\bibitem{sheng2018iterative}
H.~Sheng, J.~Chen, Y.~Zhang, W.~Ke, Z.~Xiong, and J.~Yu, ``Iterative multiple hypothesis tracking with tracklet-level association,'' \emph{IEEE Transactions on Circuits and Systems for Video Technology}, vol.~29, no.~12, pp. 3660--3672, 2018.

\end{thebibliography}

\begin{IEEEbiography}[{\includegraphics[width=1in,height=1.25in,clip,keepaspectratio]{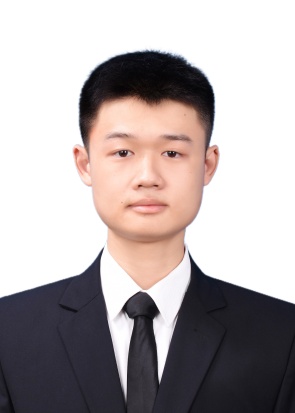}}]{Zhiwei Ning} received his B.S. degree in control science and engineering from Xi’an Jiaotong University, Xi’an, China in 2023. He is currently pursuing the Ph.D. student at the Institute of Image Processing and Pattern Recognition, Shanghai Jiao Tong University, Shanghai, China. His research interest is 3D object detection.
 \end{IEEEbiography}
 
\begin{IEEEbiography}[{\includegraphics[width=1in,height=1.25in,clip,keepaspectratio]{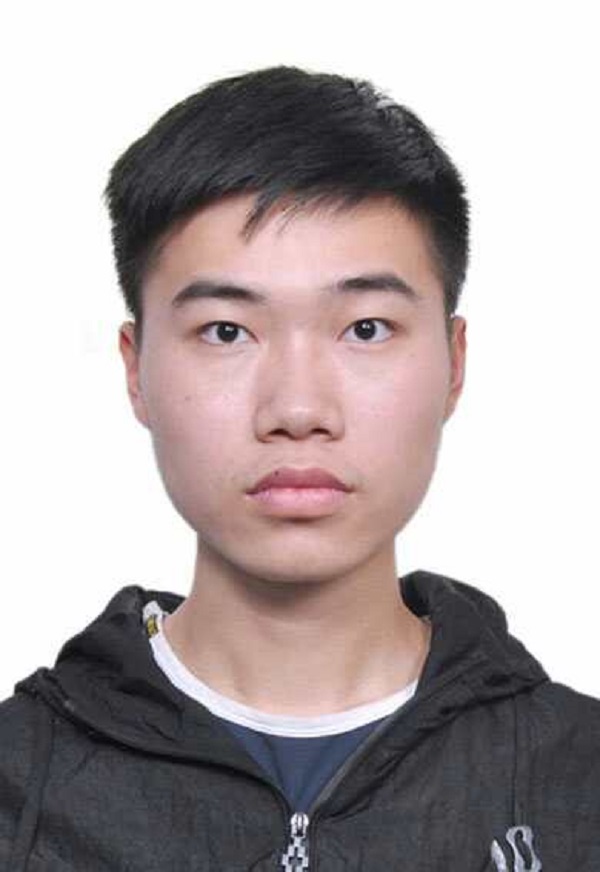}}]{Zhaojiang Liu} received his B.S. degree in the School of Automation Engineering from University of Electronic Science and Technology, Chengdu, China in 2021. He is currently pursuing the Ph.D. degree at the Institute of Image Processing and Pattern Recognition, Shanghai Jiao Tong University, Shanghai, China. His research interest is  mainly about 3D object perception.
\end{IEEEbiography}
 \vspace{-1.1 cm}

\begin{IEEEbiography}[{\includegraphics[width=1in,height=1.25in,clip,keepaspectratio]{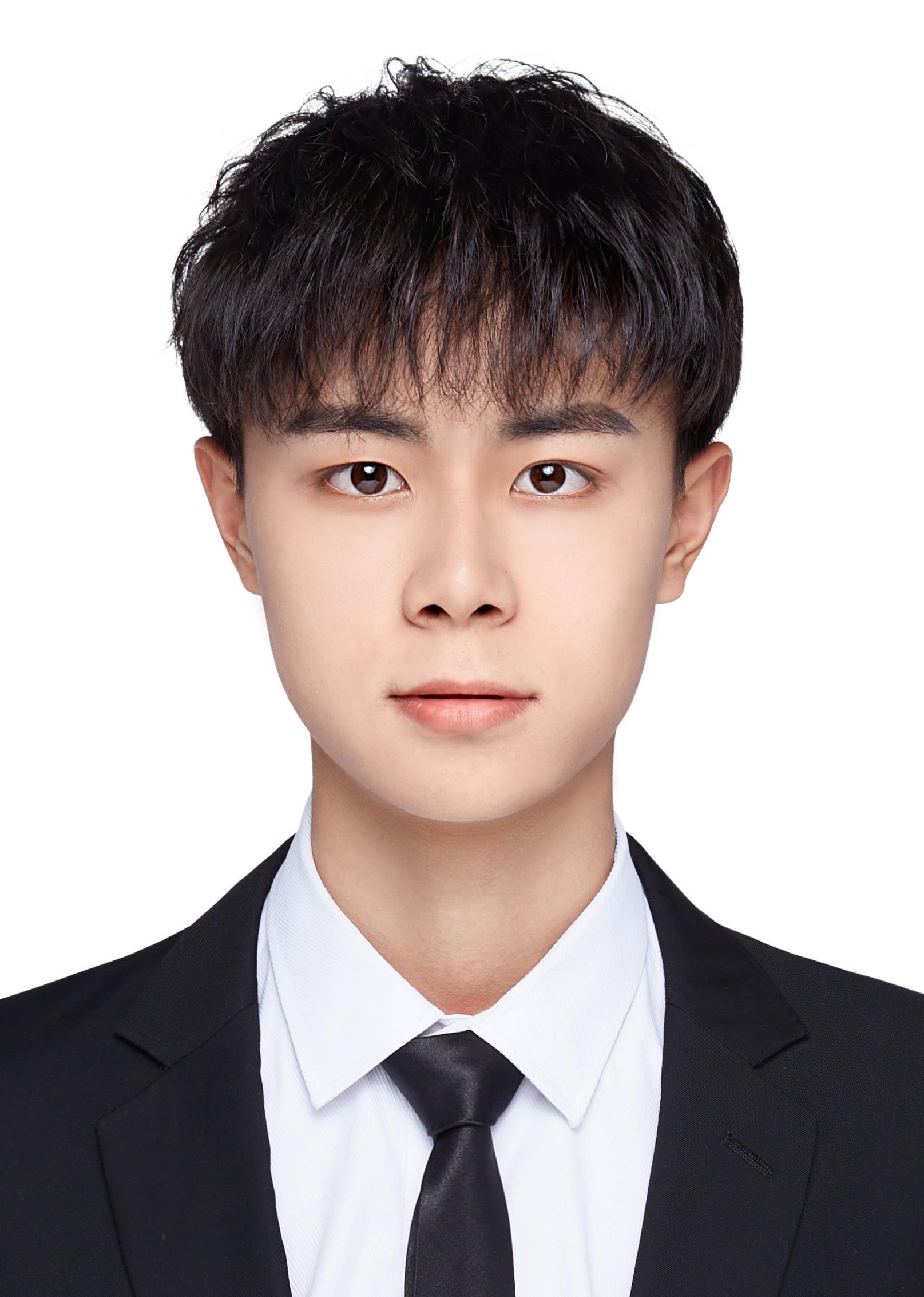}}]{Xuanang Gao} received his B.S. degree in Tianjin University, Tianjin, China in 2023. He is now a Ph.D. student at the Institute of Image Processing and Pattern Recognition, Shanghai Jiao Tong University, Shanghai, China. His research interest is self-supervised depth estimation.
 \end{IEEEbiography}
  \vspace{-1.1 cm}

\begin{IEEEbiography}[{\includegraphics[width=1in,height=1.25in,clip,keepaspectratio]{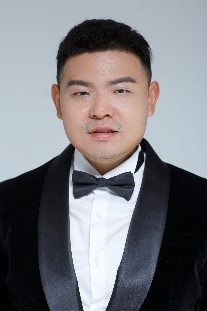}}]{Yifang Zuo} received the Ph.D. degree from the University of Technology Sydney, Ultimo, NSW, Australia, in 2018. He is currently an Associate Professor with the School of Computing and Artificial Intelligence, Jiangxi University of Finance and Economics. His research interests include Image/Point Cloud Processing. The corresponding papers have been published in major international journals such as IEEE Transactions on Image Processing, IEEE Transactions on Circuits and Systems for Video Technology, IEEE Transactions on Multimedia, and top conferences such as SIGGRAPH, AAAI.
\end{IEEEbiography}
 \vspace{-1.1 cm}

\begin{IEEEbiography}[{\includegraphics[width=1in,height=1.25in,clip,keepaspectratio]{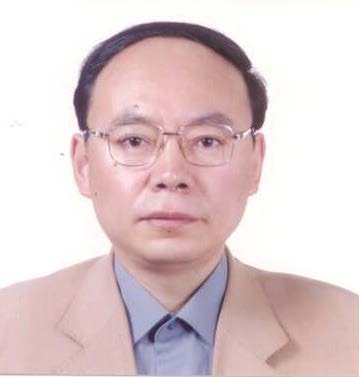}}]{Jie Yang} received his Ph.D. from the Department of Computer Science, Hamburg University, Hamburg, Germany, in 1994. Currently, he is a professor at the Institute of Image Processing and Pattern recognition, Shanghai Jiao Tong University, Shanghai, China. He has led many research projects (e.g., National Science Foundation, 863 National High Technique Plan), had one book published in Germany, and authored more than 300 journal papers. His major research interests are object detection and recognition, data fusion and data mining, and medical image processing.
\end{IEEEbiography}
 \vspace{-1.1 cm}

\begin{IEEEbiography}[{\includegraphics[width=1in,height=1.25in,clip,keepaspectratio]{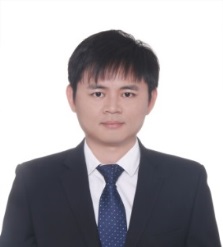}}]{Yuming Fang} (M'13-SM'17) received the B.E. degree from Sichuan University, Chengdu, China, the M.S. degree from the Beijing University of Technology, Beijing, China, and the Ph.D. degree from Nanyang Technological University, Singapore. He is currently a Professor with the School of Information Management, Jiangxi University of Finance and Economics, Nanchang, China. His research interests include visual attention modeling, visual quality assessment, computer vision, and 3D image/video processing. He serves on the editorial board for IEEE Transactions on Multimedia and Signal Processing: Image Communication.
\end{IEEEbiography}
 \vspace{-1.1 cm}

\begin{IEEEbiography}[{\includegraphics[width=1in,height=1.25in,clip,keepaspectratio]{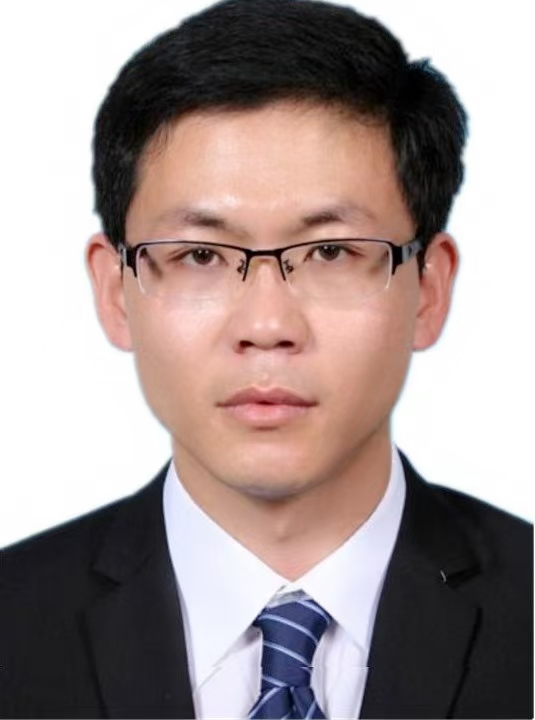}}]{Wei Liu} received the B.S. degree from Xi'an Jiaotong University, Xi'an, China, in 2012. He received the Ph.D. degree from Shanghai Jiao Tong University, Shanghai, China in 2019. He was a research fellow in The University of Adelaide and The University of Hong Kong from 2018 to 2021 and 2021 to 2022, respectively. He has been working as an associate professor in Shanghai Jiao Tong University since 2022. His current research areas include image filtering, 3D detection, 3D reconstruction and self-supervised depth estimation.
\end{IEEEbiography}
 \vspace{-1.1 cm}

\end{document}